%% file: 0-main.tex
\documentclass{article}
\PassOptionsToPackage{authoryear,compress}{natbib}
\usepackage[preprint]{neurips}
\usepackage[table]{xcolor} 
\usepackage[most]{tcolorbox}
\usepackage{longtable}
\usepackage[utf8]{inputenc} 
\usepackage[T1]{fontenc}    

\usepackage{url}            
\usepackage{booktabs}       
\usepackage{amsfonts}       
\usepackage{nicefrac}       
\usepackage{microtype}      
\usepackage{graphicx}  
\usepackage{amsmath}
\usepackage{amsthm}
\usepackage{enumitem}
\usepackage{multirow}
\usepackage{booktabs} 
\usepackage{pifont} 
\usepackage{times}
\usepackage{latexsym}
\usepackage{floatrow}

\usepackage{fancyvrb} 
\definecolor{qinglan}{RGB}{0,153,204}
\definecolor{myBlue}{RGB}{31,119,180}
\definecolor{myGreen}{RGB}{44,160, 44}
\definecolor{customGreen}{RGB}{0,176,80}
\definecolor{customBlue}{RGB}{73,198,243}
\definecolor{deepGreen}{RGB}{0,200,0}
\definecolor{myRed}{RGB}{214,39,40}

\usepackage[T1]{fontenc}

\usepackage[utf8]{inputenc}

\usepackage{microtype}

\usepackage{inconsolata}

\usepackage{amsmath}
\usepackage{graphicx}
\usepackage{amssymb}
\usepackage{longtable}
\usepackage{booktabs}
\usepackage{multirow}
\usepackage{makecell}
\usepackage{CJKutf8}
\usepackage{varwidth}
\usepackage{tikz}
\usepackage{mathrsfs}
\usepackage{ragged2e}
\usepackage{setspace}
\usepackage{caption}
\usepackage{overpic}
\usepackage{textcomp}
\usepackage{wasysym}
\usepackage[english]{babel}
\usepackage[ruled,vlined]{algorithm2e}
\usepackage{wrapfig}
\usepackage{bm}
\usepackage{arydshln}
\usepackage[inkscapelatex=false]{svg}
\usepackage{hyperref}       
\usepackage[normalem]{ulem}

\newlength{\reduce}
\setlist{nosep}
\renewcommand{\paragraph}[1]{\par\textbf{#1}\hspace{1em}}

\title{Mirage or Method? How Model–Task Alignment Induces Divergent RL Conclusions}
\author{
Haoze Wu$^1$\thanks{Equal Contribution. Work done during visit to HKUST.}
\quad Cheng Wang$^2$$^*$
\quad Wenshuo Zhao$^1$
\quad Junxian He$^3$ \\
$^1$Zhejiang University \quad
$^2$National University of Singapore \quad 
$^3$HKUST \\
\texttt{waithz@zuaa.zju.edu.cn}\quad \texttt{wangcheng@u.nus.edu}
\quad \texttt{junxianh@cse.ust.hk}
}

\begin{document}
\maketitle
 \begin{abstract}
Recent advances in applying reinforcement learning (RL) to large language models (LLMs) have led to substantial progress. In particular, a series of remarkable yet often counterintuitive phenomena have been reported in LLMs, exhibiting patterns not typically observed in traditional RL settings.
For example, notable claims include that a single training example can match the performance achieved with an entire dataset, that the reward signal does not need to be very accurate, and that training solely with negative samples can match or even surpass sophisticated reward-based methods. However, the precise conditions under which these observations hold—and, critically, when they fail—remain unclear.
In this work, we identify a key factor that differentiates RL observations: whether the pretrained model already exhibits strong \emph{Model-Task Alignment}, as measured by pass@k accuracy on the evaluated task. 
Through a systematic and comprehensive examination of a series of counterintuitive claims, supported by rigorous experimental validation across different model architectures and task domains, our findings show that while standard RL training remains consistently robust across settings, many of these counterintuitive results arise only when the model and task already exhibit strong model-task alignment. In contrast, these techniques fail to drive substantial learning in more challenging regimes, where standard RL methods remain effective. Code is available at \url{https://github.com/hkust-nlp/model-task-align-rl}.

\begin{figure}[h]
    \centering    
    \vspace{10pt}
\includegraphics[width=0.93\linewidth]{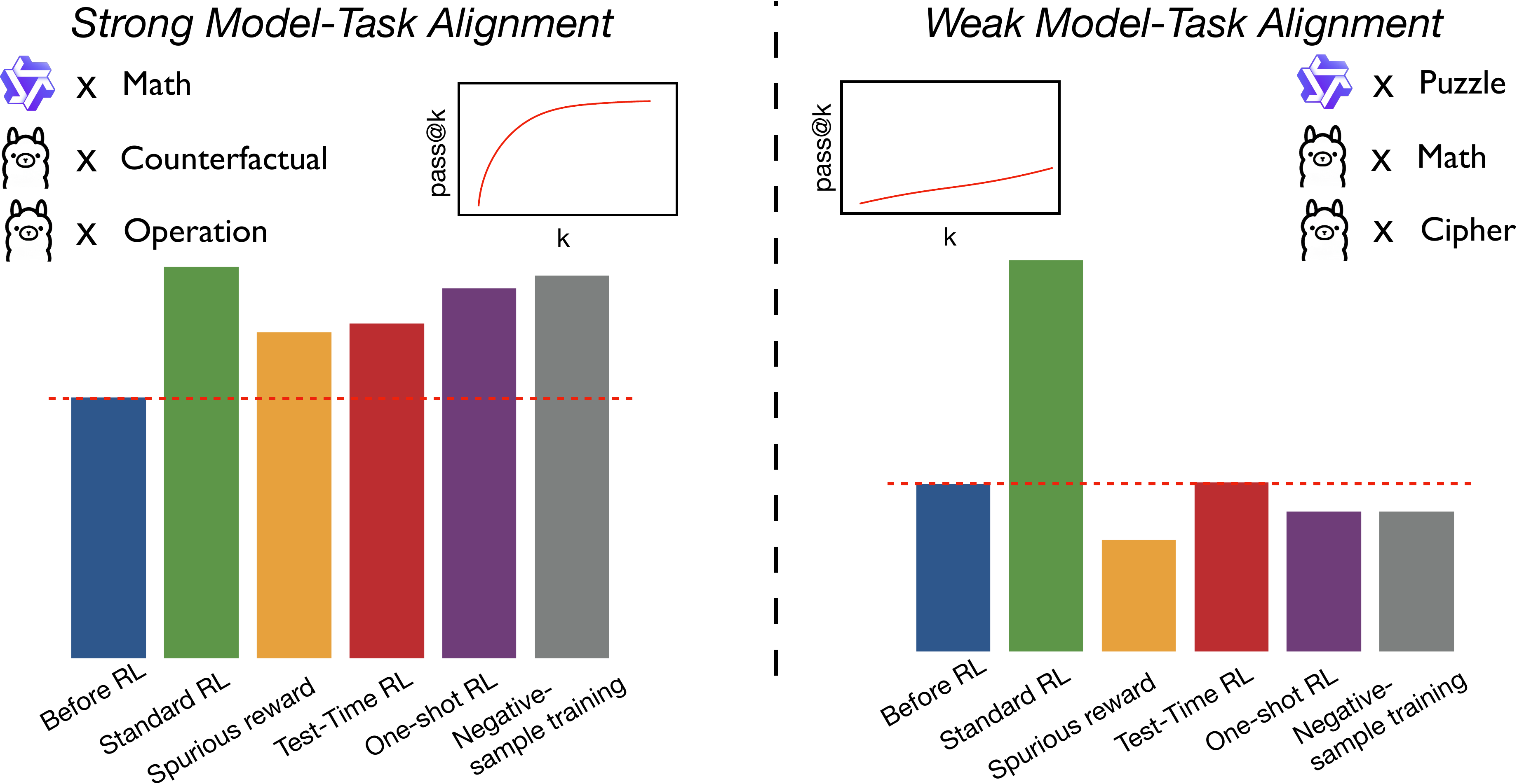}
    \caption{Model-task alignment, which is measured by pass@k accuracy on the evaluated task, drives distinct outcomes from the same series of RL approaches.}
    \label{fig:summary}
\end{figure}

\end{abstract}

\input{1-introduction}
\input{2-motivation}
\input{3-0_experiments}

\input{3-1}
\input{3-2}

\input{3-3}
\input{4-conclusion}

\bibliography{reference}
\bibliographystyle{plainnat}

\newpage
\appendix

\section{Implementation Details}
\label{app:implementation_details}
Following the setting described in Section~\ref{sec:setup}, we train with different rewards for 300 steps on mathematical and logical reasoning tasks, respectively.
The format reward is different from that of \citet{shao2025spurious}, we use the same template as SynLogic.
In addition to this, the definition of the reward functions is consistent.

\section{More Pass@k Results}
\label{app:passK}

\begin{figure}[!ht]
    \centering
    \includegraphics[width=1\linewidth]{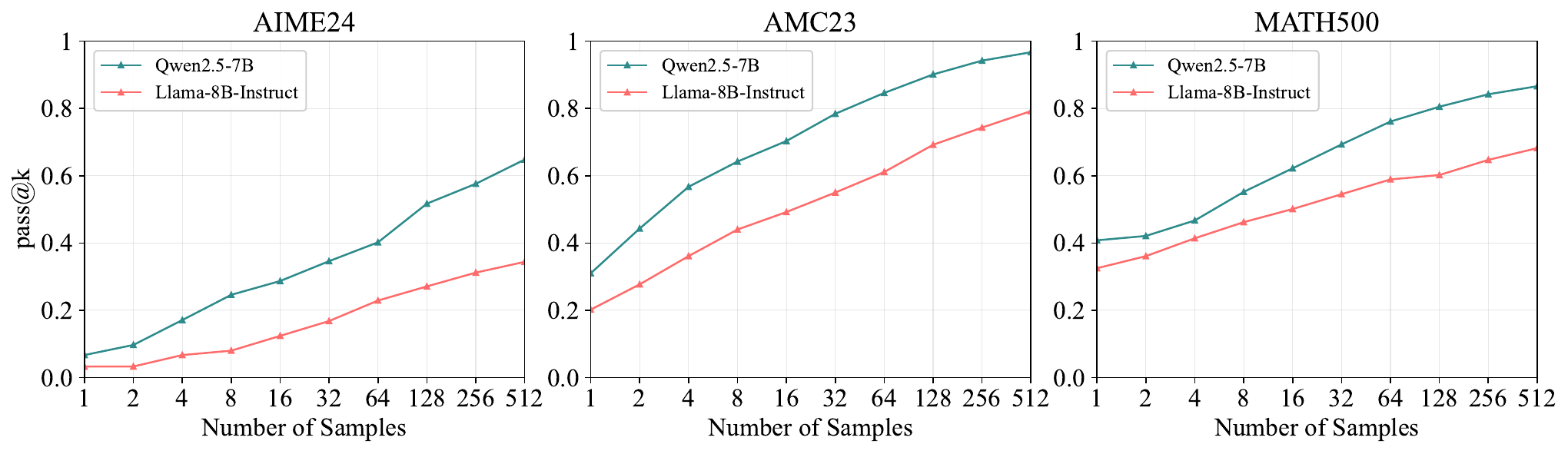}
    \caption{Pass@k for math tasks. Qwen demonstrates strong capabilities across all three mathematical evaluation datasets.}
    \label{fig:pass_k_math}
\end{figure}

\begin{figure}[!ht]
    \centering
    \includegraphics[width=1\linewidth]{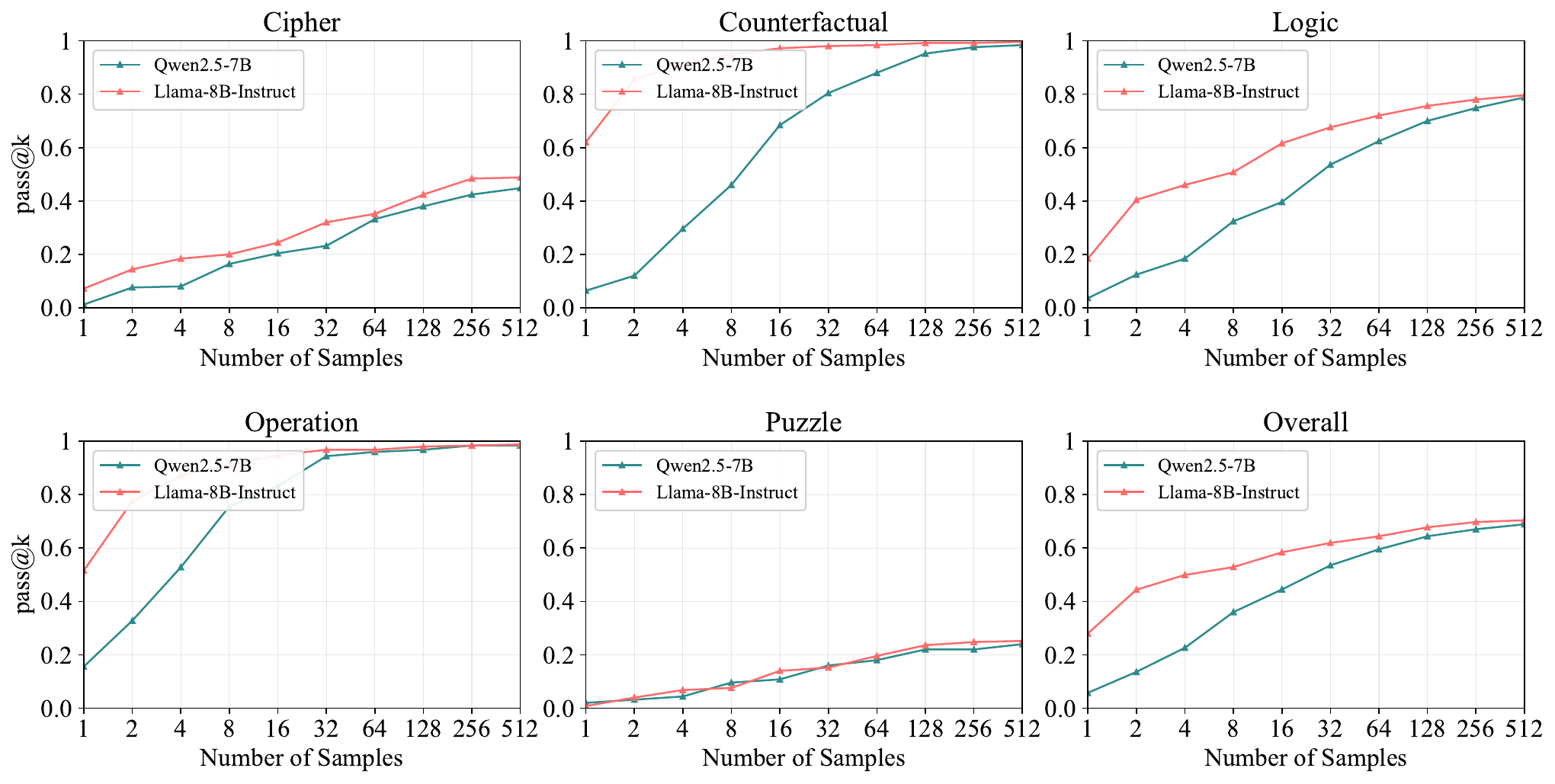}
    \caption{Pass@k for KOR-Bench. Both models demonstrate strong inherent reasoning capabilities in Operation and Counterfactual subtasks, but exhibit limited inherent logical reasoning abilities in Cipher, Puzzle and Logic.}
    \label{fig:placeholder}
\end{figure}

\section{Contamination Evaluation}
\label{app:contamination}
\subsection{Implementation Details}

Our contamination analysis follows a systematic prompt truncation methodology to evaluate potential data leakage across model-task combinations. Original prompts are truncated at varying ratios (0.4, 0.6, and 0.8) while preserving word boundaries, and models are asked to complete the remaining content using greedy decoding for deterministic outputs. We measure contamination using ROUGE-L scores between model completions and the actual remaining prompt content, where a perfect score of 1.0 indicates complete reconstruction and potential contamination. The evaluation pipeline employs distributed processing to handle complex mathematical expressions and prevent evaluation timeouts, with results aggregated across multiple rollouts to ensure statistical reliability.

\subsection{More Results}

\input{tables/more_contamination}

\subsection{Discussion about RQ1}
\label{app:discussion_RQ1}

\paragraph{How Different Reward Signals Affect the Behavior of LLMs.}
\citet{shao2025spurious} observed that in mathematical tasks, employing ground truth rewards decreases the frequency of code usage in model responses. Their study also revealed that, in contrast to Qwen2.5-Math \citep{yang2024qwen25math}, the accuracy improvement of the Qwen2.5 Base model was primarily attributed to a shift from code-based reasoning to language-based reasoning. 
As shown in Table~\ref{tab:code}, we identify analogous trends in mathematical tasks. Specifically, for logic puzzles, the application of ground truth rewards similarly reduces the incidence of code in responses. However, other types of rewards, particularly format and random rewards, do not demonstrate a significant impact on diminishing code usage frequency.
We speculate that, throughout the RL training process, ground truth rewards can steer the model away from its old reasoning pattern ( i.e., producing reasoning responses with code ) and toward a more natural, language-based reasoning pattern.

\input{tables/code}

As shown in Table~\ref{tab:diff_rewards}, spurious rewards are effective only on the Operation and Counterfactual for the Llama model; consequently, we also report the frequency of code-based reasoning before and after training on these two tasks.
As shown in Table~\ref{tab:code-llama}, we observe that, both before and after RL training, Llama almost never invokes code during the reasoning process.
We attribute the sporadic use of code (0.8) to the fact that some SynLogic tasks explicitly require outputs to be presented as code blocks.
This indicates that Llama and Qwen exhibit distinct reasoning patterns even though they both benefit from noisy reward signals in these settings.

\input{tables/code-llama}

\section{More Discussion about Difficult Example in One-shot RL}
\label{app:difficult-one-shot}

\begin{figure}[!ht]
    \centering
    \includegraphics[width=0.9\linewidth]{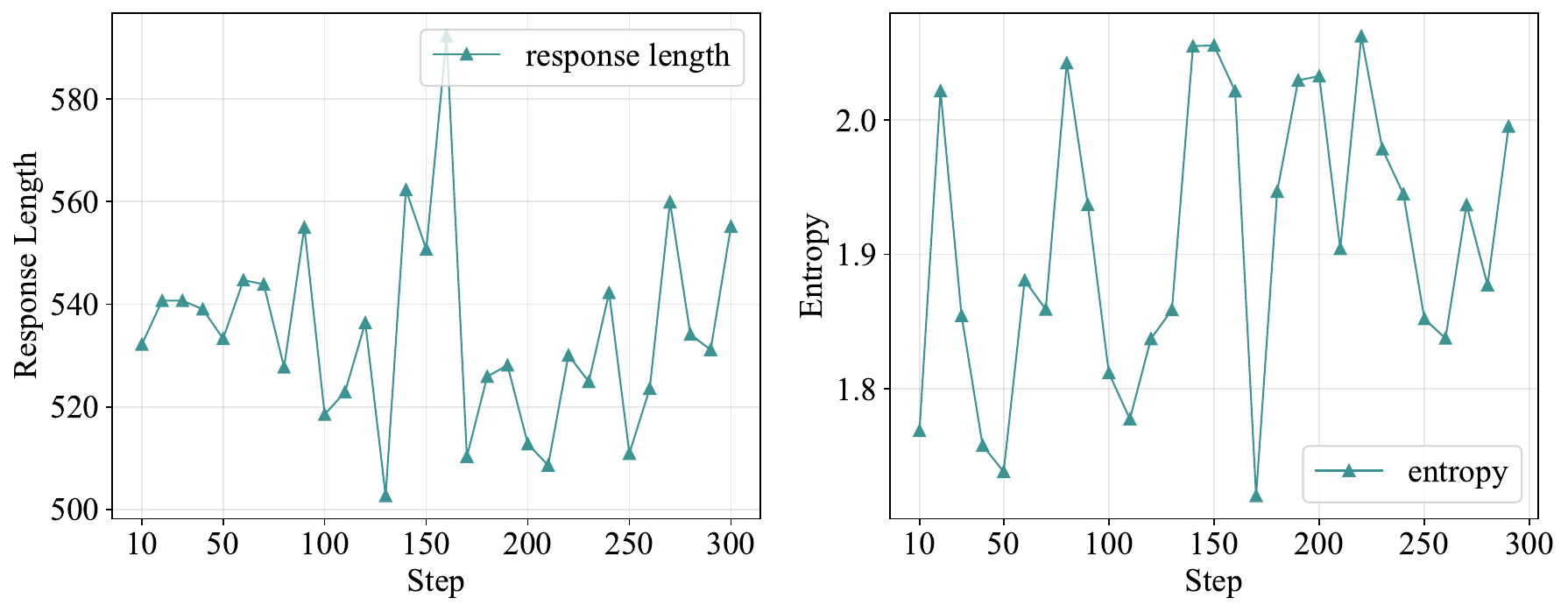}
    \caption{Training Dynamics of Qwen2.5-7B when trained with $l_{selected}$. Entropy and response length exhibit almost no changes.}
    \label{fig:difficult-1-shot}
\end{figure}

During training with $l_{selected}$, apart from the rollout accuracy (reward) remaining consistently at $0$, metrics such as entropy and response length also exhibit almost no changes.
As shown in Figure~\ref{fig:difficult-1-shot}, after 300 training steps, the model still maintains a large reinforcement learning exploration space.

\section{Few-shot RL Example Details}
\label{app:examples}

\begin{figure}[hb]
    \centering
    \begin{tcolorbox}[fonttitle=\bfseries,title=Details of example $m_{selected}$, fontlower=\small,colframe=teal,
        colback=white]
    \texttt{How many positive divisors do 9240 and 13860 have in common?}
    \end{tcolorbox}
    \caption*{}
\end{figure}

\begin{figure}[hb]
    \centering
    \begin{tcolorbox}[fonttitle=\bfseries,title=Details of example $m_{random}$, fontlower=\small,colframe=teal,
        colback=white]
    \texttt{The angles of quadrilateral $PQRS$ satisfy $\angle P = 3\angle Q = 4\angle R = 6\angle S$. What is the degree measure of $\angle P$? }
    \end{tcolorbox}
    \caption*{}
\end{figure}

\begin{figure}[hb]
    \centering
    \begin{tcolorbox}[fonttitle=\bfseries,title=Details of example $m_{random}'$, fontlower=\small,colframe=teal,
        colback=white]
    \texttt{Given a finite sequence $S=(a_1,a_2,\ldots ,a_n)$ of $n$ real numbers, let $A(S)$ be the sequence $\left(\frac{a_1+a_2}{2},\frac{a_2+a_3}{2},\ldots ,\frac{a_{n-1}+a_n}{2}\right)$ of $n-1$ real numbers. Define $A^1(S)=A(S)$ and, for each integer $m$, $2\le m\le n-1$, define $A^m(S)=A(A^{m-1}(S))$. Suppose $x>0$, and let $S=(1,x,x^2,\ldots ,x^{100})$. If $A^{100}(S)=\left(\frac{1}{2^{50}}\right)$, then what is $x$?} \textbf{AND} \texttt{If $x, 2x+2, 3x+3, \dots$ are in geometric progression, the fourth term is:}
    \end{tcolorbox}
    \caption*{}
\end{figure}

\begin{figure}[hb]
    \centering
    \begin{tcolorbox}[fonttitle=\bfseries,title=Details of example $l_{selected}$, fontlower=\small,colframe=teal,
        colback=white]
    \texttt{Here's a mathematical expression: ?-?+(6\%5)*2-?+?/?/?/4/2 = 2. The digits on the left side of the equation have been replaced with question marks. Each question mark corresponds to a digit between 0 and 9. You need to try replacing the question marks with the correct digits to restore the expression.Please put the complete expression with the filled - in digits between [[ and ]] at the end of your response, with no other content, like this: [[2 + 4 * 3 - 4 = 10]]}
    \end{tcolorbox}
    \caption*{}
\end{figure}

\begin{figure}[hb]
    \centering
    \begin{tcolorbox}[fonttitle=\bfseries,title=Details of example $l_{random}$, fontlower=\small,colframe=teal,
        colback=white]
    \texttt{Solve this cryptarithm: RRYUU + UYR + U = RYUUU (where RRYUU is a 5-digit number, UYR is a 3-digit number, U is a 1-digit number, and RYUUU is a 5-digit number). Each letter represents a unique digit. Find the digit substitution that makes the equation true.}
    \end{tcolorbox}
    \caption*{}
\end{figure}

\begin{figure}[hb]
    \centering
    \begin{tcolorbox}[fonttitle=\bfseries,title=Details of example $l_{random}'$, fontlower=\small,colframe=teal,
        colback=white]
    \texttt{In this Number Wall puzzle, add walls (marked as 'A') to divide the grid into islands. Each island must contain exactly one number, and its size must equal that number.
    \\Grid:\\+----+----+----+\\| X | 3 | X | \\+----+----+----+\\| X | X | X | \\+----+----+----+\\| X | X | X | \\+----+----+----+\\Rules:\\- Each island must contain exactly one number.\\- The total number of cells in an island (including the number cell) must equal the value of that number.\\- All cells within an island must be connected horizontally or vertically.\\- Walls (marked as 'A') cannot form 2×2 or larger continuous rectangles.\\- All islands must be separated by walls.}\\
    \textbf{AND}\\
    \texttt{In the cryptarithm: MMII + MIXIMM = MMXIIX, each letter stands for a different digit (MMII is 4 digits, MIXIMM is 6 digits, and MMXIIX is 6 digits). Determine what each letter represents to make the equation true.}
    \end{tcolorbox}
    \caption*{}
\end{figure}

\begin{figure}[hb]
    \centering
    \begin{tcolorbox}[fonttitle=\bfseries,title=Details of example $l_{simple}$, fontlower=\small,colframe=teal,
        colback=white]
    \texttt{In this word sorting challenge, you need to rearrange words in increasing based on a modified alphabet where l,z and a are the first letters. Words to sort: yachted,coelomic,harateen. Write your final answer inside: $\backslash$boxed{},like this: $\backslash$boxed{word1,word2,word3}.}
    \end{tcolorbox}
    \caption*{}
\end{figure}

\begin{figure}[hb]
    \centering
    \begin{tcolorbox}[fonttitle=\bfseries,title=Details of example $l_{mid}$, fontlower=\small,colframe=teal,
        colback=white]
    \texttt{You are an expert proficient in Dyck language, where you must complete all types of unclosed brackets (e.g., [], {}, <>) in language sequences. You need to analyze the steps of bracket pairing according to Dyck language rules. Given an initial Dyck language sequence and steps for deriving the closed bracket sequence (presented in a thinking process format), your task is to identify locations with incorrect reasoning in the Dyck language, and there may be multiple errors. This could be forgetting to close a bracket, using the wrong closing bracket, or incorrectly copying a subsequence of closing brackets in the next step. Task: Check the sequence to ensure brackets are properly closed. Input: [[()\{\}]]\{\}\\ Thought 1: We should process the input one by one and track the stack configuration.\\ Thought 2: Stack: Empty\\ Thought 3: [ ; Stack: Empty\\ Thought 4: [ ; Stack: [[ \\Thought 5: ( ; Stack: [[( \\Thought 6: ) ; Stack: [[ \\Thought 7: \{ ; Stack: [[\{ \\Thought 8: \} ; Stack: [[ \\Thought 9: ] ; Stack: [ \\Thought 10: ] ; Stack: Empty\\ Thought 11: \{ ; Stack: \{ \\Thought 12: \} ; Stack: Empty \\Thought 13: Now, we have reached the end. The final stack is empty. \\Question: Are there any reasoning errors in this sequence?}
    \end{tcolorbox}
    \caption*{}
\end{figure}

\end{document}

%% file: 1-introduction.tex
\section{Introduction}
Reinforcement Learning (RL)~\citep{sutton1998reinforcement} has emerged as a transformative post-training technique for Large Language Models (LLMs), enabling them to follow instructions~\citep{ouyang2022training} and align with human preferences~\citep{ziegler2019fine, rafailov2024directpreferenceoptimizationlanguage}. A particularly prominent application focuses on enhancing reasoning capabilities, as exemplified by breakthrough models such as OpenAI-o1~\citep{openaio1}, DeepSeek-R1~\citep{deepseekr1}, QwQ~\citep{qwq32b}, and Kimi-1.5~\citep{team2025kimi}. These systems demonstrate remarkable performance across reasoning-intensive domains including coding~\citep{jain2024livecodebenchholisticcontaminationfree}, mathematics~\citep{minervamath,he2024olympiadbench}, and logical reasoning~\citep{liu2025synlogic,chen2025enigmatascalinglogicalreasoning}.



While RL yields significant performance improvements in LLM reasoning—mirroring the success of RL in traditional domains such as games~\citep{silver2017masteringchessshogiselfplay,Silver2017MasteringT}—we also observe several remarkable yet often counterintuitive empirical phenomena. These effects appear to be unique to LLMs and would be considered unexpected in traditional RL settings.
For instance, single training examples can match or rival full-dataset training performance~\citep{wang2025reinforcement}, ground-truth reward may be surprisingly dispensable~\citep{shao2025spurious}, and training with negative samples alone can match sophisticated reward-based methods~\citep{agarwal2025unreasonableeffectivenessentropyminimization}. 

While these findings have generated considerable enthusiasm, the precise conditions under which they hold, and when they break down, remain insufficiently explored. Given that these observations may have important implications for RL practices, it is concerning that the conclusions are largely based on limited experimental settings, 
where Qwen models~\citep{qwen2.5} trained on mathematical tasks dominate the landscape.

To this end, we carry out a systematic empirical investigation of several notable RL claims, supported by rigorous experimental validation across diverse model architectures and task domains. 
Concretely, we experiment with both Qwen and non-Qwen models on math and other tasks. Our controlled experiments reveal that \emph{model-task alignment}, defined as the degree to which model capabilities match task requirements, is a critical indicator for categorizing RL observations. Specifically, models benefit from noisy rewards, test-time RL~\citep{zuo2025ttrl}, minimal training, and negative-sample training primarily within their domains of expertise, where these techniques fail for unfamiliar tasks even though standard RL training can succeed. Interestingly, we also observe that certain meta-patterns hold consistently across different settings. For instance, one-shot RL training is generally effective for the specific task to which the training example belongs, and negative-sample training helps stabilize model entropy, even though it does not always lead to overall improvements in accuracy.

We evaluate the ``alignment'' between model capabilities and task requirements using pass@k accuracy, which we find to be a reliable indicator for distinguishing these counterintuitive RL phenomena. Our hypothesis is that strong, inherent model capabilities can be readily activated through minimal training, even when guided by incorrect reward signals, whereas unfamiliar tasks demand substantially more effort—cases that we argue dominate when scaling up RL compute. Concurrent work~\citep{wu2025reasoning} investigates the mechanism behind spurious rewards and attributes their effectiveness primarily to data leakage in Qwen models on the test set. However, our results suggest otherwise: we find that spurious rewards remain effective even in the absence of contamination, provided the model already exhibits strong alignment on the evaluated task.

{Our study reveals that, unlike traditional RL training, distinct RL mechanisms emerge in the context of LLMs, depending on whether the pretrained model is already familiar with the target tasks. On the one hand, this suggests that RL phenomena should be interpreted with extra caution, as they may only reflect one of these two mechanisms. On the other hand, it also opens up opportunities for jointly optimizing base model pretraining (or mid-training) and RL post-training. For example, one might enhance the domain-specific capabilities of the base model during mid-training, enabling effective RL with limited training data and potentially inaccurate reward signals, or alternatively, allocate most compute resources to the RL stage using carefully curated training data and precise reward signals.} 

%% file: 2-motivation.tex
\section{On Unique Phenomena of RL training in LLM Reasoning}
\label{sec:mot}
Reinforcement Learning from Verifiable Rewards (RLVR) has achieved significant success in improving language model reasoning. While similar gains in accuracy from standard training have also been observed in traditional RL domains such as games, we have noticed several phenomena that appear unique to LLMs and would not typically be expected in conventional settings. For example, we highlight several remarkable, and at times counterintuitive, observations below:

\begin{itemize}[leftmargin=*]

    \item \textbf{Unexpected robustness to unreliable or absent rewards:} \citet{shao2025spurious} demonstrate that random and incorrect reward signals can improve model performance, while \citet{agarwal2025unreasonableeffectivenessentropyminimization} show that reward-free, entropy-minimization objectives can rival reward-based approaches. 
    Test-Time Reinforcement Learning (TTRL) proposed by \cite{zuo2025ttrl} further reinforces this trend by generating reward signals through aggregating majority-vote outcomes, thereby guiding the model to evolve itself on the test set.
    Together, these suggest surprising fault tolerance in RL training that challenges standard assumptions about the critical role of accurate reward signals.

    \item \textbf{One-shot training sufficiency:} \citet{wang2025reinforcement} report that training on a single carefully selected example can match or exceed performance from full dataset training, challenging assumptions about data volume requirements.

    \item \textbf{Negative-only signal effectiveness:} \citet{zhu2025surprisingeffectivenessnegativereinforcement} demonstrate that using exclusively negative reward signals achieves comparable results to standard RL training while maintaining beneficial entropy properties.
\end{itemize}

These findings carry significant implications. If broadly confirmed, they would necessitate shifts in resource allocation—such as prioritizing data selection algorithms over dataset scale, questioning the necessity of highly accurate reward modeling, and potentially introducing new research directions.
Therefore, we believe it is important to assess whether these conclusions hold in general, and if not, under what conditions they succeed or fail. Clarifying these patterns would not only help us understand the limitations of the current findings but also, in the opposite direction, reveal new opportunities for modifying models so that these findings become valid, thereby making RL training substantially easier. In this work, we will investigate these observations through controlled experiments comprehensively.

\subsection{Central Hypothesis: Model-Task Alignment Dependency}
\label{sec:model_task_alignment}

As most of the findings discussed above are based on mathematical reasoning tasks using Qwen models~\citep{qwen2.5,yang2025qwen3technicalreport}, a natural question arises: do these results generalize to other settings? For instance,~\citet{shao2025spurious} reported that spurious rewards were ineffective with Llama~\citep{llama3} models on mathematical tasks. However, we argue that treating Qwen+math as merely a special case is an overly superficial categorization. It remains unclear what specifically makes Qwen+math unique, and what the deeper, more essential factors might be.
We propose a guiding hypothesis for designing and categorizing experimental settings, which we call \textbf{Model-Task Alignment Dependency}: \emph{the effectiveness of these unique RL findings fundamentally depend on the degree of alignment between a model's inherent capabilities and the requirements of the task domain}. In other words, they depend on the model's proficiency on the evaluated task. This hypothesis may or may not hold, but we will use it as a framework to categorize experimental settings in terms of whether the model–task combination is aligned or misaligned.

\textbf{Quantifying Model-Task Alignment with pass@k.} To systematically evaluate the degree of alignment between a model's inherent capabilities and the requirements of specific task domains, we employ the pass@k metric as our primary measure of model-task proficiency. Pass@k represents the probability that at least one correct solution appears among k independent samples generated by the model for a given problem. 
This metric effectively captures how well a model's existing knowledge and reasoning patterns align with the demands of a particular task.

Formally, for a problem $x_i$ from evaluation dataset $\mathcal{D}$, we generate $n$ samples ($n \geq k$) and count the number of correct samples as $c_i$, then the unbiased estimator of pass@k over the dataset is:

$$\text{pass@k} := \mathbb{E}_{x_i \sim \mathcal{D}} \left[ 1 - \frac{\binom{n-c_i}{k}}{\binom{n}{k}} \right]$$

\begin{figure}[!t]
    \centering
\includegraphics[width=0.95\linewidth]{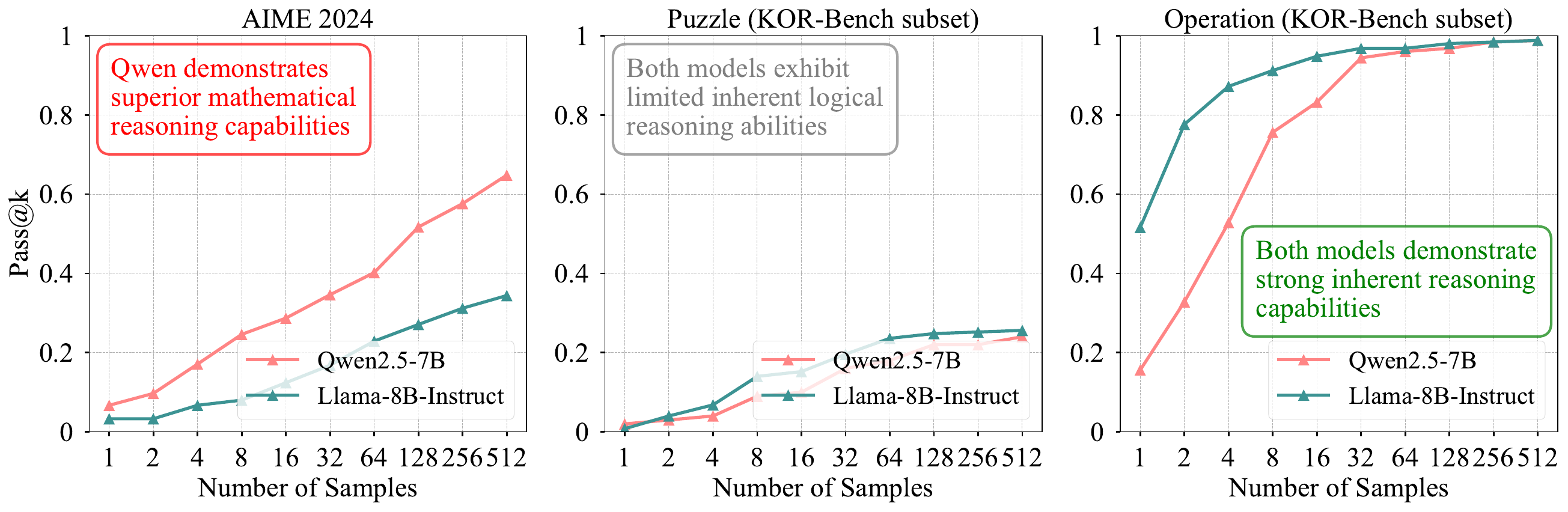}
    \caption{Pass@k for different tasks. Different LLMs have significantly different abilities on different tasks, which will affect how the RL techniques perform across model-task combinations.}
    \label{fig:pass_k}
\end{figure}

\subsection{Strategic Model and Task Selection}
\label{sec:strategic_selection}

Building on our \emph{Model-Task Alignment Dependency} hypothesis outlined in Section~\ref{sec:model_task_alignment}, we strategically design model-task combinations that test the boundaries of current claims in RL for language model reasoning. Our experimental design is motivated by the critical need to distinguish between findings that represent universal RL properties versus those that emerge from specific model-task capability alignments. We evaluate two representative language models from different families: Qwen2.5-7B-Base~\citep{qwen2.5} and Llama-3.1-8B-Instruct~\citep{llama3}, enabling systematic comparison across model architectures with varying baseline capabilities while controlling for architectural differences at comparable parameter scales.

Our evaluation encompasses mathematical and logical reasoning domains. For mathematical reasoning, we employ AIME24~\citep{aime24}, MATH500~\citep{hendrycks2021measuringmathematicalproblemsolving} and AMC23~\citep{amc23}. For logical reasoning, we utilize SynLogic~\citep{liu2025synlogic} (synthetic puzzles with 35 task types, we use the validation split), BBH~\citep{suzgun2022challenging} (multi-step reasoning tasks), BBEH~\cite{bbeh} (extended-difficulty version), and KOR-Bench~\citep{ma2024korbenchbenchmarkinglanguagemodels} (knowledge-orthogonal reasoning across five categories: Operation, Logic, Cipher, Puzzle, and Counterfactual).

To operationalize our hypothesis, we systematically measure alignment strength using pass@k metrics across all model-task combinations. As demonstrated in Figure~\ref{fig:pass_k}, models exhibit markedly different inherent capabilities across domains. Based on comprehensive evaluation (full results in Appendix~\ref{app:passK}), we identify cases of strong model-task alignment, such as Qwen2.5 on mathematical domains and both models on Operation and Counterfactual subsets of KOR-Bench, as well as weak model-task alignment cases, such as Llama3.1 on mathematical domains and both models on other logical reasoning tasks. This categorization enables us to test whether counterintuitive RL phenomena are artifacts of specific model-task alignments or represent fundamental properties of reinforcement learning in language model reasoning.

\subsection{The Contamination Hypothesis}
Concurrent work from~\citet{wu2025reasoning} proposed an alternative hypothesis, where they specifically focus on the spurious reward pattern and suggest that it stems primarily from dataset contamination during pre-training. They further confirmed the presence of data leakage in the Qwen models on several mathematical benchmarks.
While we acknowledge contamination as a valid concern, our hypothesis diverges by emphasizing the distinction between contamination and inherent task proficiency. In particular, models may demonstrate strong task performance without direct contamination of the test data.
In what follows, we categorize different experimental settings based on their contamination and inherent model-task alignment status. Later, in our experiments, we will empirically show that contamination is not the underlying cause; instead, model-task alignment serves as a more reliable differentiator.

\input{tables/contamination}
To verify our hypothesis, we extend contamination analysis beyond Qwen-Math combinations. Following~\citet{wu2025reasoning}, we evaluate model generation given partial prompts while preserving word boundaries (more details are provided in Appendix~\ref{app:contamination}). We employ greedy decoding and calculate both exact match (EM) rates and ROUGE-L scores, where ROUGE-L scores of 1.0 indicate perfect reconstruction. Table~\ref{tab:contamination} alongside Appendix~\ref{app:contamination} show that Operation and Counterfactual subsets have no contamination, yet both models demonstrate strong inherent reasoning capabilities with high pass@k scores (see Appendix~\ref{app:passK}). As we will show in our experiments, contamination is not the necessary condition for the effectiveness of these RL phenomena. Based on our contamination analysis and pass@k measurements, we categorize experimental settings into three groups:

\begin{itemize}[leftmargin=*]
   \item \textcolor{red}{\textbf{Red (Potential Contamination + Strong Model-Task Alignment)}}: Qwen2.5 on mathematical domains.
   \item \textcolor{gray}{\textbf{Gray (No Contamination + Weak Model-Task Alignment)}}: Llama3.1 on mathematical domains; both models on SynLogic, BBH, BBEH, and Logic, Cipher, Puzzle subsets of KOR-Bench.
   \item \textcolor{green!70!black}{\textbf{Green (No Contamination + Strong Model-Task Alignment)}}: Both models on Operation and Counterfactual subsets of KOR-Bench.
\end{itemize}

%% file: tables/contamination.tex
\begin{table}[!t] 
\centering 
\footnotesize 
\renewcommand{\arraystretch}{1.1} 
\setlength{\tabcolsep}{6pt} 
\scalebox{0.9}{
\begin{tabular}{l c cc cc cc cc} 
\toprule 
\multirow{2.5}{*}{\textbf{Model}} & \multirow{2.5}{*}{\textbf{Portion}} & \multicolumn{2}{c}{\textbf{AMC 23}} & \multicolumn{2}{c}{\textbf{MATH500}} & \multicolumn{2}{c}{\textbf{Puzzle}} & \multicolumn{2}{c}{\textbf{Operation}} \\ 
\cmidrule(lr){3-4} \cmidrule(lr){5-6} \cmidrule(lr){7-8} \cmidrule(lr){9-10} 
& & \textbf{ROUGE} & \textbf{EM} & \textbf{ROUGE} & \textbf{EM} & \textbf{ROUGE} & \textbf{EM} & \textbf{ROUGE} & \textbf{EM} \\ 
\midrule 
\multirow{3}{*}{\textbf{Qwen2.5-7B}} 
& 0.4 & \cellcolor{red!15}63.78 & \cellcolor{red!15}23.91 & \cellcolor{red!15}50.36 & \cellcolor{red!15}8.20 & \cellcolor{gray!15}19.56 & \cellcolor{gray!15}0.00 & \cellcolor{green!15}21.37 & \cellcolor{green!15}0.00 \\ 
& 0.6 & \cellcolor{red!15}64.42 & \cellcolor{red!15}33.73 & \cellcolor{red!15}60.98 & \cellcolor{red!15}21.20 & \cellcolor{gray!15}19.62 & \cellcolor{gray!15}0.00 & \cellcolor{green!15}24.25 & \cellcolor{green!15}0.00 \\ 
& 0.8 & \cellcolor{red!15}73.23 & \cellcolor{red!15}49.39 & \cellcolor{red!15}66.42 & \cellcolor{red!15}40.20 & \cellcolor{gray!15}19.24 & \cellcolor{gray!15}0.00 & \cellcolor{green!15}20.18 & \cellcolor{green!15}0.00 \\ 
\midrule 
\multirow{3}{*}{\textbf{Llama-3.1-8B}} 
& 0.4 & \cellcolor{gray!15}27.18 & \cellcolor{gray!15}0.00 & \cellcolor{gray!15}23.09 & \cellcolor{gray!15}0.60 & \cellcolor{gray!15}18.27 & \cellcolor{gray!15}0.00 & \cellcolor{green!15}21.83 & \cellcolor{green!15}0.00 \\ 
& 0.6 & \cellcolor{gray!15}30.64 & \cellcolor{gray!15}0.00 & \cellcolor{gray!15}40.56 & \cellcolor{gray!15}3.80 & \cellcolor{gray!15}17.31 & \cellcolor{gray!15}0.00 & \cellcolor{green!15}18.34 & \cellcolor{green!15}0.00 \\ 
& 0.8 & \cellcolor{gray!15}44.54 & \cellcolor{gray!15}4.81 & \cellcolor{gray!15}48.33 & \cellcolor{gray!15}17.8 & \cellcolor{gray!15}15.85 & \cellcolor{gray!15}0.00 & \cellcolor{green!15}16.75 & \cellcolor{green!15}0.00 \\ 
\bottomrule 
\end{tabular}}
\caption{
Contamination Analysis across model-task combinations. Portion refers to the truncation ratio of the original prompt used to test whether models can complete the remaining content.
\textcolor{red}{\textbf{Red}} \textcolor{red}{indicates potential contamination with strong model-task alignment}; 
\textcolor{gray}{\textbf{Gray}} \textcolor{gray}{indicates no contamination with weak model-task alignment};
\textcolor{green!70!black}{\textbf{Green}} \textcolor{green!70!black}{indicates no contamination with strong model-task alignment.}
}

\label{tab:contamination} 

\end{table}


%% file: 3-0_experiments.tex
\section{Experimental Setup}
\label{sec:setup}

\paragraph{Training Datasets and Evaluation.}
Except for the experiments on Test-Time RL (Section~\ref{sec:ttrl}), we use DeepScaleR~\citep{deepscaler2025} as the training set for mathematical tasks and the training split of SynLogic-Easy~\citep{liu2025synlogic} for logical tasks.
Evaluation datasets are as described in Section~\ref{sec:strategic_selection}.
Following SynLogic~\citep{liu2025synlogic}, all evaluations are conducted in a zero-shot setting, with avg@8 metrics computed for AIME 2024 and SynLogic to mitigate variance.

\paragraph{Training Configuration.}
Our experiments default to using the DAPO algorithm unless otherwise specified.
We set $\epsilon_{low}=0.2,\epsilon_{high}=0.28,\text{max promt length=2048}, \text{max generation length=8192}$.
We use dynamic sampling, and set $\text{max\_num\_gen\_batches}=2$.
During actual training, we found that for logical task training, each sampled batch often contains very few samples with non-zero reward variance.
We made two improvements: (1) When neither of the two generated sampling batches contains any samples with non-zero reward variance (which usually happens in the early stages of SynLogic training when the model cannot get any questions right), we use the second generated batch as the training batch. (2) When the number of available samples from the two generations is less than the training batch size, we duplicate the samples to match the training batch size.
We don't use length penalty.
During most training experiments, we set $lr=1e^{-6},\text{batch size}=128,\text{mini batch size}=64,\text{temperature}=1.0$.

%% file: 3-1.tex
\section{RQ1 -- Reward Signal: How Critical Is It?}
\label{sec:reward}

This section investigates the role of reward signal quality and its impact on RL performance for LLMs. Previous work in Reinforcement Learning with Human Feedback has shown that more accurate reward models do not always lead to better downstream performance \citep{chen2024accuracy}. In the specific context of LLM reasoning, initial studies found that models with strong inherent reasoning abilities exhibit surprising robustness to noisy reward signals, whereas weaker models show poor noise tolerance \citep{lv2025climb, shao2025spurious}. 
Building on these findings, we extend the analysis by considering diverse reward signals across different model-task combinations. Implementation details are presented in Appendix~\ref{app:implementation_details}.

\input{tables/diff_reward}

\subsection{Results}
We present results in Table~\ref{tab:diff_rewards}.
Based on the experimental results, we identify three critical findings regarding the impact of reward signal quality on model performance (Appendix~\ref{app:discussion_RQ1} provides additional discussion):

\paragraph{Ground Truth Rewards Consistently Outperform All Alternatives.}
Across both model families and all task domains, utilizing ground truth rewards consistently yields the highest performance improvements. For instance, Qwen2.5-7B achieves substantial gains on AIME24 ($14.2$ vs. baseline $3.3$) and MATH500 ($71.0$ vs. baseline $40.8$) when trained with accurate reward signals. This establishes ground truth rewards as the gold standard for RL training in reasoning tasks.

\paragraph{Model-Task Alignment Determines Robustness to Noisy Rewards.}
The effectiveness of noisy reward signals depends on model-task alignment strength across our three experimental categories. In settings with strong alignment (\textcolor{red}{Red} and \textcolor{green!70!black}{Green} categories), models demonstrate surprising robustness to spurious rewards, with Qwen2.5-7B maintaining reasonable performance on mathematical tasks and both models showing improvements on Operation and Counterfactual tasks even with random rewards. Conversely, in \textcolor{gray}{weak alignment} settings, spurious rewards consistently fail to provide meaningful improvements, as seen with Llama3.1-8B on mathematical tasks and both models on challenging logical reasoning benchmarks. This pattern confirms that alignment strength, rather than contamination alone, determines robustness to noisy rewards.

\paragraph{Limited Effectiveness of Self-Rewarded Methods.}
Self-Rewarded Reinforcement Learning methods, including majority voting and entropy minimization, consistently underperform compared to external reward-based approaches. While self-rewarded methods shows some promise on mathematical tasks for Qwen2.5-7B (majority vote achieves $69.4$ on MATH500), it fails to match the performance of ground truth external rewards and shows poor generalization to logical reasoning tasks across both model families.

\subsection{Test-Time RL}
\label{sec:ttrl}

Test-Time Reinforcement Learning (TTRL)~\citep{zuo2025ttrl} addresses a fundamental challenge in LLM development: how to improve model performance on unlabeled test data without access to ground-truth labels for reward signals.
It prompts the model to generate multiple responses to each test question and use the most frequent answer as the label for reward signals.
Although the model is trained on the unlabeled test set, this approach is essentially no different from Self-Rewarded Reinforcement Learning when majority voting is employed.
Thus, we are also curious whether TTRL remains effective for different models and in domains beyond mathematics.

Table~\ref{tab:ttrl} shows the results of the Qwen and Llama models on different tasks.
Due to the limited scale of the test dataset, we trained for 30 steps on all test datasets.
It could be observed that in settings where the model–task alignment is strong, TTRL yields substantial improvements, as exemplified by Qwen on math tasks and Operation subset.
For tasks in which the model lacks initial prior knowledge, TTRL fails to deliver improvements or yields only marginal gains.

\input{tables/ttrl}

As discussed by~\citet{zuo2025ttrl}, majority voting is the foundation of TTRL. 
We also recorded the variation of Maj@16 during the training process; the results are shown in Table~\ref{tab:maj}.
We can observe that, in settings where TTRL yields substantial improvements, Maj@16 consistently rises throughout training.
Especially for Qwen on Operation subset, it achieves an absolute gain of 16.4 points.
This further underscores that TTRL’s efficacy hinges on strong model–task alignment, rather than on contamination.

\input{tables/maj}

%% file: tables/diff_reward.tex
\begin{table}[!t]
\centering
\scriptsize
\setlength{\tabcolsep}{3pt}
\renewcommand{\arraystretch}{1.2}
\scalebox{0.93}{
\begin{tabular}{@{}l*{11}{c}@{}}
\toprule
& \multicolumn{3}{c}{\textbf{Math Tasks}} & \multicolumn{8}{c}{\textbf{Logic Tasks}} \\
\cmidrule(lr){2-4} \cmidrule(lr){5-12}
& \multirow{2.5}{*}{\textbf{AIME24}}
& \multirow{2.5}{*}{\textbf{MATH500}}
& \multirow{2.5}{*}{\textbf{AMC}}
& \multirow{2.5}{*}{\textbf{SynLogic}}
& \multirow{2.5}{*}{\textbf{BBH}}
& \multirow{2.5}{*}{\textbf{BBEH}}
& \multicolumn{5}{c}{\textbf{KOR Benchmark}} \\
\cmidrule(lr){8-12}
& & & & & & &
\textbf{OP} &
\textbf{CF} &
\textbf{Puzzle} &
\textbf{Logic} &
\textbf{Cipher} \\
\midrule
\midrule
\multicolumn{12}{c}{\cellcolor{white}\textbf{Qwen2.5-7B Family}} \\
\cellcolor{white}Base & \cellcolor{red!15}3.3 & \cellcolor{red!15}40.8 & \cellcolor{red!15}31.0 & \cellcolor{gray!15}1.5 & \cellcolor{gray!15}45.2 & \cellcolor{gray!15}1.2 & \cellcolor{green!15}27.2 & \cellcolor{green!15}17.2 & \cellcolor{gray!15}0.8 & \cellcolor{gray!15}8.0 & \cellcolor{gray!15}4.8 \\
\midrule
\multicolumn{12}{c}{\cellcolor{white}RLVR (External Reward)} \\
\cellcolor{white}Correct & \cellcolor{red!15}14.2\textcolor{myGreen}{$_{+10.9}$} & \cellcolor{red!15}71.0\textcolor{myGreen}{$_{+30.2}$} & \cellcolor{red!15}62.4\textcolor{myGreen}{$_{+31.4}$} & \cellcolor{gray!15}42.6\textcolor{myGreen}{$_{+41.1}$} & \cellcolor{gray!15}62.7\textcolor{myGreen}{$_{+17.5}$} & \cellcolor{gray!15}6.8\textcolor{myGreen}{$_{+5.6}$} & \cellcolor{green!15}82.4\textcolor{myGreen}{$_{+55.2}$} & \cellcolor{green!15}79.6\textcolor{myGreen}{$_{+62.4}$} & \cellcolor{gray!15}16.8\textcolor{myGreen}{$_{+10.0}$} & \cellcolor{gray!15}46.4\textcolor{myGreen}{$_{+38.4}$} & \cellcolor{gray!15}20.4\textcolor{myGreen}{$_{+15.6}$} \\
\cellcolor{white}Random & \cellcolor{red!15}10.0\textcolor{myGreen}{$_{+6.7}$} & \cellcolor{red!15}57.5\textcolor{myGreen}{$_{+16.7}$} & \cellcolor{red!15}45.7\textcolor{myGreen}{$_{+14.7}$} & \cellcolor{gray!15}10.2\textcolor{myGreen}{$_{+8.7}$} & \cellcolor{gray!15}32.7\textcolor{myRed}{$_{-12.5}$} & \cellcolor{gray!15}0.0\textcolor{myRed}{$_{-1.2}$} & \cellcolor{green!15}53.6\textcolor{myGreen}{$_{+26.4}$} & \cellcolor{green!15}30.8\textcolor{myGreen}{$_{+13.6}$} & \cellcolor{gray!15}1.2\textcolor{myGreen}{$_{+0.4}$} & \cellcolor{gray!15}6.8\textcolor{myRed}{$_{-1.2}$} & \cellcolor{gray!15}3.6\textcolor{myRed}{$_{-1.2}$} \\
\cellcolor{white}Incorrect & \cellcolor{red!15}6.7\textcolor{myGreen}{$_{+3.4}$} & \cellcolor{red!15}57.0\textcolor{myGreen}{$_{+16.2}$} & \cellcolor{red!15}43.1\textcolor{myGreen}{$_{+12.1}$} & \cellcolor{gray!15}0.0\textcolor{myRed}{$_{-1.5}$} & \cellcolor{gray!15}30.3\textcolor{myRed}{$_{-14.9}$} & \cellcolor{gray!15}0.0\textcolor{myRed}{$_{-1.2}$} & \cellcolor{green!15}60.8\textcolor{myGreen}{$_{+33.6}$} & \cellcolor{green!15}12.8\textcolor{myRed}{$_{-4.4}$} & \cellcolor{gray!15}0.4\textcolor{myRed}{$_{-0.4}$} & \cellcolor{gray!15}6.4\textcolor{myRed}{$_{-1.6}$} & \cellcolor{gray!15}3.2\textcolor{myRed}{$_{-1.6}$} \\
\cellcolor{white}Format & \cellcolor{red!15}6.7\textcolor{myGreen}{$_{+3.4}$} & \cellcolor{red!15}55.3\textcolor{myGreen}{$_{+14.5}$} & \cellcolor{red!15}48.9\textcolor{myGreen}{$_{+17.9}$} & \cellcolor{gray!15}1.5\textcolor{myRed}{$_{0.0}$} & \cellcolor{gray!15}44.4\textcolor{myRed}{$_{-0.8}$} & \cellcolor{gray!15}2.4\textcolor{myGreen}{$_{+1.2}$} & \cellcolor{green!15}37.2\textcolor{myGreen}{$_{+10.0}$} & \cellcolor{green!15}21.6\textcolor{myGreen}{$_{+4.4}$} & \cellcolor{gray!15}0.8\textcolor{myRed}{$_{0.0}$} & \cellcolor{gray!15}6.8\textcolor{myRed}{$_{-1.2}$} & \cellcolor{gray!15}4.4\textcolor{myRed}{$_{-0.4}$} \\
\midrule
\multicolumn{12}{c}{\cellcolor{white}Self-Rewarded Reinforcement Learning} \\
\cellcolor{white}Vote & \cellcolor{red!15}13.3\textcolor{myGreen}{$_{+10.0}$} & \cellcolor{red!15}69.4\textcolor{myGreen}{$_{+28.6}$} & \cellcolor{red!15}58.2\textcolor{myGreen}{$_{+27.2}$} & \cellcolor{gray!15}2.8\textcolor{myGreen}{$_{+1.3}$} & \cellcolor{gray!15}33.6\textcolor{myRed}{$_{-11.6}$} & \cellcolor{gray!15}0.0\textcolor{myRed}{$_{-1.2}$} & \cellcolor{green!15}56.4\textcolor{myGreen}{$_{+29.2}$} & \cellcolor{green!15}16.3\textcolor{myRed}{$_{-0.9}$} & \cellcolor{gray!15}0.8\textcolor{myRed}{$_{0.0}$} & \cellcolor{gray!15}6.8\textcolor{myRed}{$_{-1.2}$} & \cellcolor{gray!15}3.2\textcolor{myRed}{$_{-1.6}$} \\
\cellcolor{white}EM & \cellcolor{red!15}11.6\textcolor{myGreen}{$_{+8.3}$} & \cellcolor{red!15}70.8\textcolor{myGreen}{$_{+30.0}$} & \cellcolor{red!15}57.8\textcolor{myGreen}{$_{+26.8}$} & \cellcolor{gray!15}1.5\textcolor{myRed}{$_{0.0}$} & \cellcolor{gray!15}37.5\textcolor{myRed}{$_{-7.7}$} & \cellcolor{gray!15}0.0\textcolor{myRed}{$_{-1.2}$} & \cellcolor{green!15}67.2\textcolor{myGreen}{$_{+40.0}$} & \cellcolor{green!15}27.2\textcolor{myGreen}{$_{+10.0}$} & \cellcolor{gray!15}0.8\textcolor{myRed}{$_{0.0}$} & \cellcolor{gray!15}6.8\textcolor{myRed}{$_{-37.6}$} & \cellcolor{gray!15}3.2\textcolor{myRed}{$_{-13.6}$} \\
\midrule
\midrule
\multicolumn{12}{c}{\cellcolor{white}\textbf{Llama3.1-8B-Instruct Family}} \\
\cellcolor{white}Base & \cellcolor{gray!15}3.3 & \cellcolor{gray!15}32.5 & \cellcolor{gray!15}20.2 & \cellcolor{gray!15}0.8 & \cellcolor{gray!15}38.6 & \cellcolor{gray!15}4.1 & \cellcolor{green!15}60.4 & \cellcolor{green!15}86.4 & \cellcolor{gray!15}2.0 & \cellcolor{gray!15}28.8 & \cellcolor{gray!15}8.4 \\
\midrule
\multicolumn{12}{c}{\cellcolor{white}RLVR (External Reward)} \\
\cellcolor{white}Correct & \cellcolor{gray!15}6.7\textcolor{myGreen}{$_{+3.4}$} & \cellcolor{gray!15}38.6\textcolor{myGreen}{$_{+6.1}$} & \cellcolor{gray!15}25.1\textcolor{myGreen}{$_{+4.9}$} & \cellcolor{gray!15}21.0\textcolor{myGreen}{$_{+20.2}$} & \cellcolor{gray!15}49.1\textcolor{myGreen}{$_{+10.5}$} & \cellcolor{gray!15}4.3\textcolor{myGreen}{$_{+0.2}$} & \cellcolor{green!15}76.0\textcolor{myGreen}{$_{+15.6}$} & \cellcolor{green!15}88.8\textcolor{myGreen}{$_{+2.4}$} & \cellcolor{gray!15}15.6\textcolor{myGreen}{$_{+13.6}$} & \cellcolor{gray!15}34.4\textcolor{myGreen}{$_{+7.6}$} & \cellcolor{gray!15}11.6\textcolor{myGreen}{$_{+3.2}$} \\
\cellcolor{white}Random & \cellcolor{gray!15}3.3\textcolor{myRed}{$_{0.0}$} & \cellcolor{gray!15}26.8\textcolor{myRed}{$_{-5.7}$} & \cellcolor{gray!15}21.3\textcolor{myGreen}{$_{+1.1}$} & \cellcolor{gray!15}0.0\textcolor{myRed}{$_{-0.8}$} & \cellcolor{gray!15}32.1\textcolor{myRed}{$_{-6.5}$} & \cellcolor{gray!15}4.1\textcolor{myRed}{$_{0.0}$} & \cellcolor{green!15}69.2\textcolor{myGreen}{$_{+8.8}$} & \cellcolor{green!15}87.2\textcolor{myGreen}{$_{+0.8}$} & \cellcolor{gray!15}0.8\textcolor{myRed}{$_{-1.2}$} & \cellcolor{gray!15}23.6\textcolor{myRed}{$_{-5.2}$} & \cellcolor{gray!15}4.4\textcolor{myRed}{$_{-4.0}$}\\
\cellcolor{white}Incorrect & \cellcolor{gray!15}2.1\textcolor{myRed}{$_{-1.2}$} & \cellcolor{gray!15}26.4\textcolor{myRed}{$_{-6.1}$} & \cellcolor{gray!15}18.7\textcolor{myRed}{$_{-1.5}$} & \cellcolor{gray!15}0.8\textcolor{myRed}{$_{0.0}$} & \cellcolor{gray!15}30.2\textcolor{myRed}{$_{-8.4}$} & \cellcolor{gray!15}3.8\textcolor{myRed}{$_{-0.3}$} & \cellcolor{green!15}70.0\textcolor{myGreen}{$_{+9.6}$} & \cellcolor{green!15}83.2\textcolor{myRed}{$_{-3.2}$} & \cellcolor{gray!15}0.8\textcolor{myRed}{$_{-1.2}$} & \cellcolor{gray!15}19.2\textcolor{myRed}{$_{-9.6}$} & \cellcolor{gray!15}4.4\textcolor{myRed}{$_{-4.0}$} \\
\cellcolor{white}Format & \cellcolor{gray!15}3.1\textcolor{myRed}{$_{-0.2}$} & \cellcolor{gray!15}31.5\textcolor{myRed}{$_{-1.0}$} & \cellcolor{gray!15}18.7\textcolor{myRed}{$_{-1.5}$} & \cellcolor{gray!15}0.8\textcolor{myRed}{$_{0.0}$} & \cellcolor{gray!15}36.4\textcolor{myRed}{$_{-2.2}$} & \cellcolor{gray!15}4.1\textcolor{myRed}{$_{0.0}$} & \cellcolor{green!15}68.8\textcolor{myGreen}{$_{+8.4}$} & \cellcolor{green!15}85.6\textcolor{myRed}{$_{-0.8}$} & \cellcolor{gray!15}2.0\textcolor{myRed}{$_{0.0}$} & \cellcolor{gray!15}28.0\textcolor{myRed}{$_{-0.8}$} & \cellcolor{gray!15}6.4\textcolor{myRed}{$_{-2.0}$} \\
\midrule
\multicolumn{12}{c}{\cellcolor{white}Self-Rewarded Reinforcement Learning} \\
\cellcolor{white}Vote & \cellcolor{gray!15}4.6\textcolor{myGreen}{$_{+1.3}$} & \cellcolor{gray!15}37.7\textcolor{myGreen}{$_{+5.2}$} & \cellcolor{gray!15}23.0\textcolor{myGreen}{$_{+2.8}$} & \cellcolor{gray!15}1.5\textcolor{myGreen}{$_{+0.7}$} & \cellcolor{gray!15}35.9\textcolor{myRed}{$_{-2.7}$} & \cellcolor{gray!15}4.3\textcolor{myGreen}{$_{+0.2}$} & \cellcolor{green!15}67.2\textcolor{myGreen}{$_{+6.8}$} & \cellcolor{green!15}83.2\textcolor{myRed}{$_{-3.2}$} & \cellcolor{gray!15}2.0\textcolor{myRed}{$_{0.0}$} & \cellcolor{gray!15}28.0\textcolor{myRed}{$_{-0.8}$} & \cellcolor{gray!15}8.8\textcolor{myGreen}{$_{+0.4}$} \\
\cellcolor{white}EM & \cellcolor{gray!15}5.1\textcolor{myGreen}{$_{+1.8}$} & \cellcolor{gray!15}38.3\textcolor{myGreen}{$_{+5.8}$} & \cellcolor{gray!15}25.0\textcolor{myGreen}{$_{+4.8}$} & \cellcolor{gray!15}0.8\textcolor{myRed}{$_{0.0}$} & \cellcolor{gray!15}34.8\textcolor{myRed}{$_{-3.8}$} & \cellcolor{gray!15}4.1\textcolor{myRed}{$_{0.0}$} & \cellcolor{green!15}73.6\textcolor{myGreen}{$_{+13.2}$} & \cellcolor{green!15}87.2\textcolor{myGreen}{$_{+0.8}$} & \cellcolor{gray!15}2.0\textcolor{myRed}{$_{0.0}$} & \cellcolor{gray!15}23.6\textcolor{myRed}{$_{-5.2}$} & \cellcolor{gray!15}7.6\textcolor{myRed}{$_{-0.8}$} \\
\bottomrule
\end{tabular}}
\caption{Comprehensive evaluation of different reward signals in RL. ``Vote'' denotes Majority Voting, ``EM'' means entropy minimization on self-generated samples only; OP: Operation ; CF: Counterfactual. 
\textcolor{red}{\textbf{Red}} \textcolor{red}{indicates potential contamination with strong model-task alignment}; 
\textcolor{gray}{\textbf{Gray}} \textcolor{gray}{indicates no contamination with weak model-task alignment};
\textcolor{green!70!black}{\textbf{Green}} \textcolor{green!70!black}{indicates no contamination with strong model-task alignment.}
}
\label{tab:diff_rewards}
\end{table}

%% file: tables/ttrl.tex
\begin{table}[!t]
    \centering
    \scalebox{0.8}{
    \begin{tabular}{lccc|lccc}
        \toprule
        \textbf{Model} & \textbf{MATH500} & \textbf{SynLogic} & \textbf{OP} & \textbf{Model} & \textbf{MATH500} & \textbf{SynLogic} & \textbf{OP} \\
        \midrule
        Qwen2.5-7B & \cellcolor{red!15}40.8 & \cellcolor{gray!15}1.5 & \cellcolor{green!15}27.2 & Llama-3.1-8B-Instruct & \cellcolor{gray!15}32.5 & \cellcolor{gray!15}0.8 & \cellcolor{green!15}60.4 \\
        +TTRL & \cellcolor{red!15}62.1\textcolor{myGreen}{\(_{+21.3}\)} & \cellcolor{gray!15}1.8\textcolor{myGreen}{\(_{+0.3}\)} & \cellcolor{green!15}55.6\textcolor{myGreen}{\(_{+28.4}\)} & +TTRL & \cellcolor{gray!15}41.2\textcolor{myGreen}{\(_{+8.7}\)} & \cellcolor{gray!15}0.8\textcolor{myRed}{\(_{0.0}\)} & \cellcolor{green!15}83.6\textcolor{myGreen}{\(_{+23.2}\)} \\
        \bottomrule
    \end{tabular}}
    \caption{Test-Time Reinforcement Learning (TTRL) performance changes. TTRL produces significant gains only when model-task alignment is strong (\textcolor{red}{red} and \textcolor{green!70!black}{green} cells).}
    \label{tab:ttrl}
\end{table}

%% file: tables/maj.tex
\begin{table}[!t]
\centering
\small
\begin{tabular}{l|ccccccc}
\toprule
 & Step 0 & Step 5 & Step 10 & Step 15 & Step 20 & Step 25 & Step 30 \\ 
\midrule
Qwen+Math500 & \cellcolor{red!15}54.2 & \cellcolor{red!15}60.6 & \cellcolor{red!15}64.3 & \cellcolor{red!15}68.2 & \cellcolor{red!15}67.1 & \cellcolor{red!15}69.3 & \cellcolor{red!15}70.5\small{$+16.3$} \\ 
Qwen+SynLogic & \cellcolor{gray!15}2.2 & \cellcolor{gray!15}3.0 & \cellcolor{gray!15}3.7 & \cellcolor{gray!15}4.4 & \cellcolor{gray!15}4.4 & \cellcolor{gray!15}4.4 & \cellcolor{gray!15}5.2\small{$+3.0$} \\ 
Qwen+OP & \cellcolor{green!15}46.0 & \cellcolor{green!15}53.6 & \cellcolor{green!15}55.6 & \cellcolor{green!15}57.2 & \cellcolor{green!15}58.8 & \cellcolor{green!15}60.0 & \cellcolor{green!15}60.0\small{$+16.4$} \\ 
\midrule
Llama+Math500 & \cellcolor{gray!15}46.3 & \cellcolor{gray!15}48.6 & \cellcolor{gray!15}51.3 & \cellcolor{gray!15}53.2 & \cellcolor{gray!15}53.9 & \cellcolor{gray!15}55.0 & \cellcolor{gray!15}54.7\small{$+{8.4}$} \\
Llama+SynLogic & \cellcolor{gray!15}1.5 & \cellcolor{gray!15}1.5 & \cellcolor{gray!15}2.2 & \cellcolor{gray!15}1.5 & \cellcolor{gray!15}2.2 & \cellcolor{gray!15}2.2 & \cellcolor{gray!15}2.2\small{$+0.7$} \\ 
Llama+OP & \cellcolor{green!15}73.6 & \cellcolor{green!15}78.0 & \cellcolor{green!15}79.6 & \cellcolor{green!15}84.0 & \cellcolor{green!15}83.6 & \cellcolor{green!15}86.8 & \cellcolor{green!15}88.4\small{$+14.8$} \\ 
\bottomrule
\end{tabular}
\caption{
The variation of Maj@16 as training progresses. In tasks where TTRL brings significant improvements (\textcolor{red}{red} and \textcolor{green!70!black}{green}), Maj@16 continues to improve with training.}
\label{tab:maj}
\end{table}

%% file: 3-2.tex
\section{RQ2 -- Is One-shot Enough for RL to Work?}
\label{sec:one-shot}
\citet{wang2025reinforcement} demonstrated that training on a single carefully selected question can yield performance comparable to full dataset training, challenging conventional assumptions about data volume requirements in RL. 
\citet{wang2025reinforcement} designs a selection algorithm based on the variance of training rewards, and we denote samples selected by this algorithm as $m_{selected}$ for mathematical tasks and $l_{selected}$ for logical tasks. In addition to that, we also randomly selected one or two samples from the dataset to form $(m_{random}, l_{random})$ and $(m'_{random}, l'_{random})$ for comparison. The specific examples we used are detailed in Appendix~\ref{app:examples}. The remaining experimental settings are consistent with those described in Section~\ref{sec:setup}, and we train models for 300 steps.

\input{tables/one-shot}

\subsection{Results}
We present results in Table~\ref{tab:one-shot}. Based on the experimental results, we identify two critical findings regarding the effectiveness of one-shot reinforcement learning:

\textbf{One-shot RL Success Depends on Model-Task Alignment.} The effectiveness of one-shot reinforcement learning is highly contingent on the alignment between model capabilities and task domain requirements. In strong alignment settings (\textcolor{red}{Red} and \textcolor{green!70!black}{Green} categories), both models demonstrate remarkable ability to generalize from single examples: Qwen2.5-7B achieves performance comparable to full dataset training on mathematical tasks (MATH500: $65.2$ vs. full training $71.0$), while both Qwen2.5-7B and Llama3.1-8B-Instruct show substantial improvements on Operation and Counterfactual tasks (e.g., Llama on Operation: $69.2$ vs. baseline $60.4$). However, this success does not extend to \textcolor{gray}{weak alignment} settings, where both models show minimal improvements across challenging logical reasoning benchmarks. This suggests that one-shot RL serves as an effective fine-tuning mechanism only when models already possess strong foundational capabilities in the target domain.

\textbf{Sample Selection Strategy Shows Limited Impact.} Contrary to expectations, the sophisticated sample selection algorithm proposed by~\citet{wang2025reinforcement}. does not consistently outperform random sample selection. For Qwen2.5-7B on mathematical tasks, both selected and random samples achieve similar performance levels (MATH500: selected $65.2$ vs. random $58.7$ and $63.0$), while for Llama3.1-8B-Instruct, the differences are negligible across all benchmarks. This finding challenges the assumption that reward variance-based selection provides substantial advantages over simpler random sampling approaches.

\begin{figure}[!t]
    \centering
    \includegraphics[width=0.9\linewidth]{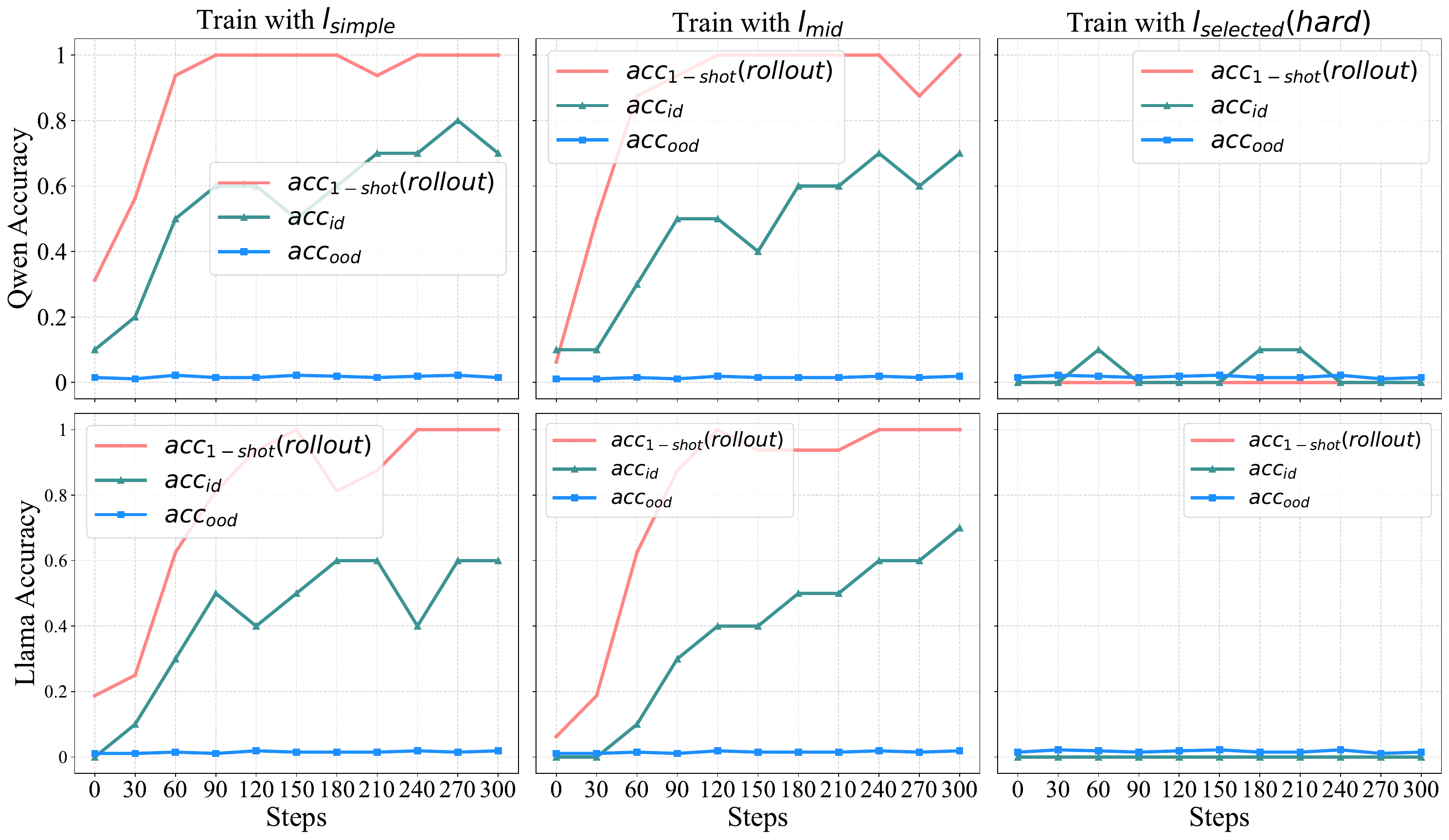}
    \caption{The changes in two models' accuracy during the training. If the initial rollout accuracy is non-zero, both models rapidly fit the employed samples ($l_{simple}, l_{mid}$) and exhibit generalization within the same subtask; however, we observe no generalization to puzzles of other types.
    }
    \label{fig:one-shot-discuss}
\end{figure}

\subsection{Discussion}

\citet{wang2025reinforcement} showed that training on a single sample for mathematical tasks can quickly improve the accuracy of that sample and also lead to improvements on the test set. We attempt to verify this conclusion on logical tasks. Considering that the initial rollout accuracy of the model on $l_{selected}$ is $0$, we additionally sample two examples whose initial rollout accuracies on Qwen2.5-7B are $5/16$ and $1/16$ (on Llama-3.1-8B-Instruct are $3/16$ and $1/16)$, denoted as $l_{simple}$ and $l_{mid}$. During training, we track three metrics: the rollout accuracy of these examples $acc_{1-shot}$, the accuracy of the subtask to which this example belongs (in-distribution) $acc_{id}$, and the accuracy of other subtasks in SynLogic (out-of-distribution) $acc_{ood}$. The results are shown in Figure~\ref{fig:one-shot-discuss}.

\textbf{One-shot RL possesses the ability to generalize within the distribution.} When the problem is relatively simple (with an initial rollout accuracy that is not zero), the model's rollout accuracy on that sample quickly increases. Although the initial rollout accuracy of $l_{mid}$ on Qwen is only one-fifth that of $l_{simple}$ (on Llama is one-third), it still attains a high rollout accuracy within a few dozen steps.
Since GRPO and DAPO compute advantages via intra-group normalization, the model is unable to derive any informative feedback from samples whose initial rollout accuracy is zero.
Moreover, we observe that the test accuracy for the same subtask also continues to improve, demonstrating effective within-distribution generalization.

\textbf{One-shot RL struggles to generalize to other types of logic puzzles.} We find that while models can improve on tasks similar to their training example, they fail to transfer learning to different puzzle types. This suggests that one-shot learning primarily exploits existing model capabilities rather than developing new reasoning skills.


%% file: tables/one-shot.tex
\begin{table}[!t]
\centering
\scriptsize
\setlength{\tabcolsep}{3pt}
\renewcommand{\arraystretch}{1.2}
\scalebox{0.93}{
\begin{tabular}{@{}c*{11}{c}@{}}
\toprule
\multirow{3.5}{*}{\textbf{Dataset}} & \multicolumn{3}{c}{\thead{\textbf{Math Tasks}}} &  \multicolumn{8}{c}{\thead{\textbf{Logic Tasks}}} \\
 \cmidrule(lr){2-4}  \cmidrule(lr){5-12}
& \multirow{2.5}{*}{\textbf{AIME24}}
& \multirow{2.5}{*}{\textbf{MATH500}}
& \multirow{2.5}{*}{\textbf{AMC}}
& \multirow{2.5}{*}{\textbf{SynLogic}}
& \multirow{2.5}{*}{\textbf{BBH}}
& \multirow{2.5}{*}{\textbf{BBEH}}
& \multicolumn{5}{c}{\textbf{KOR Benchmark}} \\
\cmidrule(lr){8-12}
& & & & & & &
\textbf{OP} &
\textbf{CF} &
\textbf{Puzzle} &
\textbf{Logic} &
\textbf{Cipher} \\
\midrule
\midrule
\multicolumn{12}{c}{\textbf{Qwen2.5-7B}} \\
\midrule
$\emptyset$ & \cellcolor{red!15}3.3 & \cellcolor{red!15}40.8 & \cellcolor{red!15}31.0  & \cellcolor{gray!15}1.5 & \cellcolor{gray!15}45.2 & \cellcolor{gray!15}1.2  & \cellcolor{green!15}27.2 & \cellcolor{green!15}17.2 & \cellcolor{gray!15}0.8 & \cellcolor{gray!15}8.0 & \cellcolor{gray!15}4.8\\
full set & \cellcolor{red!15}14.2\textcolor{myGreen}{$_{+10.9}$} & \cellcolor{red!15}71.0\textcolor{myGreen}{$_{+30.2}$} & \cellcolor{red!15}62.4\textcolor{myGreen}{$_{+31.4}$} & \cellcolor{gray!15}42.6\textcolor{myGreen}{$_{+41.1}$} & \cellcolor{gray!15}62.7\textcolor{myGreen}{$_{+17.5}$} & \cellcolor{gray!15}6.8\textcolor{myGreen}{$_{+5.6}$} & \cellcolor{green!15}82.4\textcolor{myGreen}{$_{+55.2}$} & \cellcolor{green!15}79.6\textcolor{myGreen}{$_{+62.4}$} & \cellcolor{gray!15}16.8\textcolor{myGreen}{$_{+10.0}$} & \cellcolor{gray!15}46.4\textcolor{myGreen}{$_{+38.4}$} & \cellcolor{gray!15}20.4\textcolor{myGreen}{$_{+15.6}$}\\
random-1 & \cellcolor{red!15}10.7\textcolor{myGreen}{$_{+7.4}$} & \cellcolor{red!15}58.7\textcolor{myGreen}{$_{+17.9}$} & \cellcolor{red!15}53.1\textcolor{myGreen}{$_{+22.1}$} & \cellcolor{gray!15}0.8\textcolor{myRed}{$_{-0.7}$} & \cellcolor{gray!15}40.2\textcolor{myRed}{$_{-5.0}$} & \cellcolor{gray!15}0.0\textcolor{myRed}{$_{-1.2}$} & \cellcolor{green!15}60.4\textcolor{myGreen}{$_{+33.2}$} & \cellcolor{green!15}36.8\textcolor{myGreen}{$_{+19.6}$} & \cellcolor{gray!15}0.8\textcolor{myRed}{$_{0.0}$} & \cellcolor{gray!15}6.4\textcolor{myRed}{$_{-1.6}$} & \cellcolor{gray!15}4.4\textcolor{myRed}{$_{-0.4}$}\\
random-2 & \cellcolor{red!15}12.5\textcolor{myGreen}{$_{+9.2}$}  & \cellcolor{red!15}63.0\textcolor{myGreen}{$_{+22.2}$} & \cellcolor{red!15}55.7\textcolor{myGreen}{$_{+22.7}$} & \cellcolor{gray!15}2.4\textcolor{myGreen}{$_{+0.9}$} & \cellcolor{gray!15}43.1\textcolor{myRed}{$_{-2.1}$} & \cellcolor{gray!15}1.2\textcolor{myRed}{$_{0.0}$} & \cellcolor{green!15}67.2\textcolor{myGreen}{$_{+40.0}$} & \cellcolor{green!15}56.8\textcolor{myGreen}{$_{+39.6}$} & \cellcolor{gray!15}2.0\textcolor{myGreen}{$_{+1.2}$} & \cellcolor{gray!15}3.2\textcolor{myRed}{$_{-4.8}$} & \cellcolor{gray!15}4.8\textcolor{myRed}{$_{0.0}$}\\
selected-1 & \cellcolor{red!15}12.3\textcolor{myGreen}{$_{+9.0}$} & \cellcolor{red!15}65.2\textcolor{myGreen}{$_{+24.4}$} & \cellcolor{red!15}55.2\textcolor{myGreen}{$_{+24.2}$} & \cellcolor{gray!15}0.8\textcolor{myRed}{$_{-0.7}$} & \cellcolor{gray!15}39.9\textcolor{myRed}{$_{-5.3}$} & \cellcolor{gray!15}0.0\textcolor{myRed}{$_{-1.2}$} & \cellcolor{green!15}69.2\textcolor{myGreen}{$_{+42.0}$} & \cellcolor{green!15}38.4\textcolor{myGreen}{$_{+21.2}$} & \cellcolor{gray!15}0.8\textcolor{myRed}{$_{0.0}$} & \cellcolor{gray!15}8.0\textcolor{myRed}{$_{0.0}$} & \cellcolor{gray!15}6.4\textcolor{myGreen}{$_{+1.6}$} \\

\midrule
\midrule
\multicolumn{12}{c}{\textbf{Llama3.1-8B-Instruct}} \\
\midrule
$\emptyset$ & \cellcolor{gray!15}3.3 & \cellcolor{gray!15}32.5 & \cellcolor{gray!15}20.2  & \cellcolor{gray!15}0.8 & \cellcolor{gray!15}38.6 & \cellcolor{gray!15}4.1 & \cellcolor{green!15}60.4 & \cellcolor{green!15}86.4 & \cellcolor{gray!15}2.0 & \cellcolor{gray!15}28.8 & \cellcolor{gray!15}8.4 \\
full set & \cellcolor{gray!15}6.7\textcolor{myGreen}{$_{+3.4}$} & \cellcolor{gray!15}38.6\textcolor{myGreen}{$_{+6.1}$} & \cellcolor{gray!15}25.1\textcolor{myGreen}{$_{+4.9}$} & \cellcolor{gray!15}21.0\textcolor{myGreen}{$_{+20.2}$} & \cellcolor{gray!15}49.1\textcolor{myGreen}{$_{+10.5}$} & \cellcolor{gray!15}4.3\textcolor{myGreen}{$_{+0.2}$} & \cellcolor{green!15}76.0\textcolor{myGreen}{$_{+15.6}$} & \cellcolor{green!15}88.8\textcolor{myGreen}{$_{+2.4}$} & \cellcolor{gray!15}15.6\textcolor{myGreen}{$_{+13.6}$} & \cellcolor{gray!15}34.4\textcolor{myGreen}{$_{+7.6}$} & \cellcolor{gray!15}11.6\textcolor{myGreen}{$_{+3.2}$} \\
random-1 & \cellcolor{gray!15}3.8\textcolor{myGreen}{$_{+0.5}$} & \cellcolor{gray!15}30.5\textcolor{myRed}{$_{-2.0}$} & \cellcolor{gray!15}21.1\textcolor{myGreen}{$_{+0.9}$}& \cellcolor{gray!15}0.8\textcolor{myRed}{$_{0.0}$} & \cellcolor{gray!15}35.1\textcolor{myRed}{$_{-3.5}$} & \cellcolor{gray!15}3.8\textcolor{myRed}{$_{-0.3}$} & \cellcolor{green!15}73.6\textcolor{myGreen}{$_{+13.2}$} & \cellcolor{green!15}85.6\textcolor{myRed}{$_{-0.8}$} & \cellcolor{gray!15}1.2\textcolor{myRed}{$_{-0.8}$} & \cellcolor{gray!15}28.0\textcolor{myRed}{$_{-0.8}$} & \cellcolor{gray!15}8.8\textcolor{myGreen}{$_{+0.4}$} \\
random-2 & \cellcolor{gray!15}2.7\textcolor{myRed}{$_{-0.6}$} & \cellcolor{gray!15}33.1\textcolor{myGreen}{$_{+0.6}$} & \cellcolor{gray!15}21.1\textcolor{myGreen}{$_{+0.9}$} & \cellcolor{gray!15}0.8\textcolor{myRed}{$_{0.0}$} & \cellcolor{gray!15}36.7\textcolor{myRed}{$_{-1.9}$} & \cellcolor{gray!15}4.1\textcolor{myRed}{$_{0.0}$} & \cellcolor{green!15}70.0\textcolor{myGreen}{$_{+9.6}$} & \cellcolor{green!15}86.4\textcolor{myRed}{$_{0.0}$} & \cellcolor{gray!15}2.8\textcolor{myGreen}{$_{+0.8}$} & \cellcolor{gray!15}27.2\textcolor{myRed}{$_{-1.6}$} & \cellcolor{gray!15}8.4\textcolor{myRed}{$_{0.0}$} \\
selected-1 & \cellcolor{gray!15}3.7\textcolor{myGreen}{$_{+0.4}$} & \cellcolor{gray!15}30.3\textcolor{myRed}{$_{-2.2}$} & \cellcolor{gray!15}22.3\textcolor{myGreen}{$_{+2.1}$}& \cellcolor{gray!15}0.8\textcolor{myRed}{$_{0.0}$} & \cellcolor{gray!15}34.4\textcolor{myRed}{$_{-4.2}$} & \cellcolor{gray!15}3.8\textcolor{myRed}{$_{-0.3}$} & \cellcolor{green!15}69.2\textcolor{myGreen}{$_{+8.8}$} & \cellcolor{green!15}88.8\textcolor{myGreen}{$_{+2.4}$} & \cellcolor{gray!15}2.0\textcolor{myRed}{$_{0.0}$} & \cellcolor{gray!15}19.2\textcolor{myRed}{$_{-9.6}$} & \cellcolor{gray!15}6.8\textcolor{myRed}{$_{-1.6}$} \\
\bottomrule
\end{tabular}}
\caption{One-shot Reinforcement Learning Results. OP: Operation; CF: Counterfactual. We only observe the effectiveness of one-shot reinforcement learning in settings with strong model-task alignment (\textcolor{red}{red} and \textcolor{green!70!black}{green}).}
\label{tab:one-shot}
\end{table}

%% file: 3-3.tex
\section{RQ3 — Does RL Work with Only Negative Samples?}
\label{sec:posneg}
Recent work~\citep{zhu2025surprisingeffectivenessnegativereinforcement} has demonstrated that training exclusively on negative samples can be surprisingly effective for language model reasoning. However, these findings have primarily been observed in scenarios with strong model-task alignment. We investigate whether negative-only training generalizes to weak model-task alignment scenarios, where models lack strong foundational capabilities in the target domain.

\paragraph{Implementation Details.}
In our implementation, negative Sample Reinforcement (NSR) masks out all trajectories with reward $1$ (correct answers) when computing the policy gradient, leaving only negative-rewarded samples to drive updates. Conversely, Positive Sample Reinforcement (PSR) ignores trajectories with reward $0$ and optimizes only on positively rewarded samples. All other hyperparameters remain identical to the DAPO baseline described in Section~\ref{sec:setup}, with each model trained for $300$ steps.

\subsection{Results}
\label{sec:posneg_results}

\input{tables/negative}

Table~\ref{tab:nsr} summarizes the performance of NSR and PSR relative to the full-signal DAPO baseline across our three experimental categories. The results reveal distinct patterns based on model-task alignment strength:

\textbf{Strong Model-Task Alignment Enables Effective Negative-Sample Learning.}
In settings with strong model-task alignment (\textcolor{red}{Red} and \textcolor{green}{Green} categories), both NSR and PSR show comparable effectiveness, recovering most of the performance gains achieved by full-signal DAPO. For Qwen2.5-7B on mathematical tasks, both approaches achieve \textasciitilde95\% of the DAPO improvement (MATH500: NSR 68.7 and PSR 70.3 vs. DAPO 71.0). This demonstrates that when models already possess strong domain capabilities, either positive-only or negative-only signals can effectively drive learning.

\textbf{Weak Model-Task Alignment Reveals Superior Performance of Positive-Only Signals.}
In weak alignment settings (\textcolor{gray}{Gray} category), PSR consistently outperforms NSR across logical reasoning tasks. For instance, on SynLogic, PSR enables meaningful improvements (Qwen2.5-7B: 1.5 vs. 24.8, Llama3.1-8B: 0.8 vs. 13.0), while NSR shows minimal gains. Overall, while PSR and NSR demonstrate comparable effectiveness in strong alignment settings, PSR emerges as the more robust approach in challenging domains where models lack expertise.

\begin{figure}[!t]
    \centering
    \includegraphics[width=0.98\linewidth]{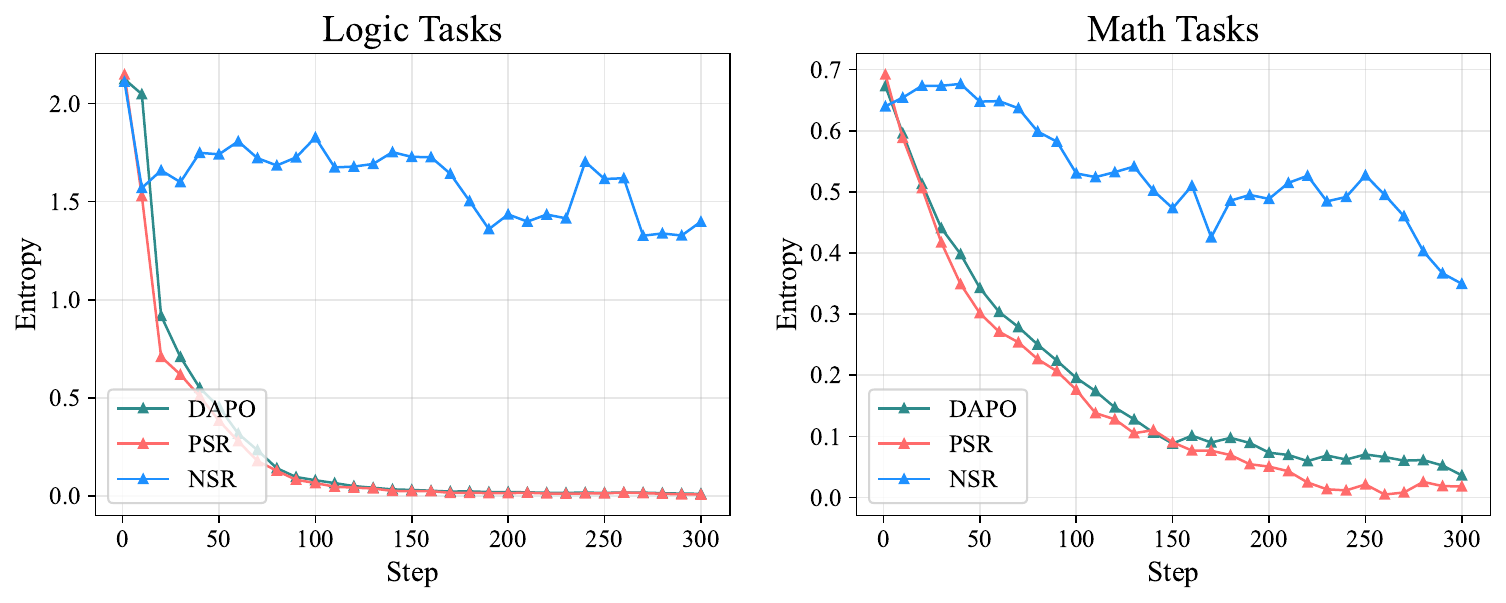}
    \caption{Entropy Dynamics of Qwen2.5-7B during Training. NSR can maintain the exploration space of reinforcement learning, but a larger exploration space is not always favorable, as in logical tasks.}
    \label{fig:entropy}
\end{figure}

\subsection{Discussion}
\label{sec:posneg_discuss}
The relationship between positive and negative samples in reinforcement learning is fundamentally connected to the exploration-exploitation trade-off, with entropy serving as a key mediator. To elucidate these dynamics in our experimental context, we examine how different sample types affect the exploration-exploitation balance through their impact on training entropy.

\paragraph{Negative Signals Help Maintain Exploration.}
Figure~\ref{fig:entropy} plots token-level entropy throughout training. Consistent with \citet{zhu2025surprisingeffectivenessnegativereinforcement}, NSR slows entropy collapse, especially on mathematical tasks—suggesting that penalising only erroneous trajectories can preserve output diversity. However, the flatter entropy curve on logical tasks corresponds to poorer final accuracy.


%% file: tables/negative.tex
\begin{table}[!ht]
\centering
\scriptsize
\setlength{\tabcolsep}{3pt}
\renewcommand{\arraystretch}{1.2}
\scalebox{0.915}{
\begin{tabular}{@{}l*{11}{c}@{}}
\toprule
& \multicolumn{3}{c}{\textbf{Math Tasks}} & \multicolumn{8}{c}{\textbf{Logic Tasks}} \\
\cmidrule(lr){2-4} \cmidrule(lr){5-12}
& \multirow{2.5}{*}{\textbf{AIME24}}
& \multirow{2.5}{*}{\textbf{MATH500}}
& \multirow{2.5}{*}{\textbf{AMC}}
& \multirow{2.5}{*}{\textbf{SynLogic}}
& \multirow{2.5}{*}{\textbf{BBH}}
& \multirow{2.5}{*}{\textbf{BBEH}}
& \multicolumn{5}{c}{\textbf{KOR Benchmark}} \\
\cmidrule(lr){8-12}
& & & & & & &
\textbf{OP} &
\textbf{CF} &
\textbf{Puzzle} &
\textbf{Logic} &
\textbf{Cipher} \\
\midrule
\midrule
\cellcolor{white}Qwen2.5-7B & \cellcolor{red!15}3.3 & \cellcolor{red!15}40.8 & \cellcolor{red!15}31.0 & \cellcolor{gray!15}1.5 & \cellcolor{gray!15}45.2 & \cellcolor{gray!15}1.2 & \cellcolor{green!15}27.2 & \cellcolor{green!15}17.2 & \cellcolor{gray!15}0.8 & \cellcolor{gray!15}8.0 & \cellcolor{gray!15}4.8 \\
\midrule
DAPO & \cellcolor{red!15}14.2\textcolor{myGreen}{$_{+10.9}$} & \cellcolor{red!15}71.0\textcolor{myGreen}{$_{+30.2}$} & \cellcolor{red!15}62.4\textcolor{myGreen}{$_{+31.4}$} & \cellcolor{gray!15}42.6\textcolor{myGreen}{$_{+41.1}$} & \cellcolor{gray!15}62.7\textcolor{myGreen}{$_{+17.5}$} & \cellcolor{gray!15}6.8\textcolor{myGreen}{$_{+5.6}$} & \cellcolor{green!15}82.4\textcolor{myGreen}{$_{+55.2}$} & \cellcolor{green!15}79.6\textcolor{myGreen}{$_{+62.4}$} & \cellcolor{gray!15}16.8\textcolor{myGreen}{$_{+10.0}$} & \cellcolor{gray!15}46.4\textcolor{myGreen}{$_{+38.4}$} & \cellcolor{gray!15}20.4\textcolor{myGreen}{$_{+15.6}$} \\
NSR & \cellcolor{red!15}13.9\textcolor{myGreen}{$_{+10.6}$} & \cellcolor{red!15}68.7\textcolor{myGreen}{$_{+27.9}$} & \cellcolor{red!15}63.5\textcolor{myGreen}{$_{+32.5}$} & \cellcolor{gray!15}1.5\textcolor{myRed}{$_{0.0}$} & \cellcolor{gray!15}41.2\textcolor{myRed}{$_{-4.0}$} & \cellcolor{gray!15}1.6\textcolor{myGreen}{$_{+0.4}$} & \cellcolor{green!15}60.4\textcolor{myGreen}{$_{+33.2}$} & \cellcolor{green!15}36.8\textcolor{myGreen}{$_{+19.6}$} & \cellcolor{gray!15}2.0\textcolor{myGreen}{$_{+1.2}$} & \cellcolor{gray!15}6.8\textcolor{myRed}{$_{-1.2}$} & \cellcolor{gray!15}4.8\textcolor{myRed}{$_{0.0}$} \\
PSR & \cellcolor{red!15}14.0\textcolor{myGreen}{$_{+10.7}$} & \cellcolor{red!15}70.3\textcolor{myGreen}{$_{+29.5}$} & \cellcolor{red!15}63.1\textcolor{myGreen}{$_{+32.1}$} & \cellcolor{gray!15}24.8\textcolor{myGreen}{$_{+23.3}$} & \cellcolor{gray!15}57.1\textcolor{myGreen}{$_{+11.9}$} & \cellcolor{gray!15}4.3\textcolor{myGreen}{$_{+3.1}$} & \cellcolor{green!15}73.6\textcolor{myGreen}{$_{+46.4}$} & \cellcolor{green!15}38.4\textcolor{myGreen}{$_{+21.2}$} & \cellcolor{gray!15}9.2\textcolor{myGreen}{$_{+8.4}$} & \cellcolor{gray!15}31.2\textcolor{myGreen}{$_{+23.2}$} & \cellcolor{gray!15}11.2\textcolor{myGreen}{$_{+6.4}$} \\
\midrule
\midrule
Llama3.1-8B & \cellcolor{gray!15}3.3 & \cellcolor{gray!15}32.5 & \cellcolor{gray!15}20.2 & \cellcolor{gray!15}0.8 & \cellcolor{gray!15}38.6 & \cellcolor{gray!15}4.1 & \cellcolor{green!15}60.4 & \cellcolor{green!15}86.4 & \cellcolor{gray!15}2.0 & \cellcolor{gray!15}28.8 & \cellcolor{gray!15}8.4  \\ 
\midrule
DAPO & \cellcolor{gray!15}6.7\textcolor{myGreen}{$_{+3.4}$} & \cellcolor{gray!15}38.6\textcolor{myGreen}{$_{+6.1}$} & \cellcolor{gray!15}25.1\textcolor{myGreen}{$_{+4.9}$} & \cellcolor{gray!15}21.0\textcolor{myGreen}{$_{+20.2}$} & \cellcolor{gray!15}49.1\textcolor{myGreen}{$_{+10.5}$} & \cellcolor{gray!15}4.3\textcolor{myGreen}{$_{+0.2}$} & \cellcolor{green!15}76.0\textcolor{myGreen}{$_{+15.6}$} & \cellcolor{green!15}88.8\textcolor{myGreen}{$_{+2.4}$} & \cellcolor{gray!15}15.6\textcolor{myGreen}{$_{+13.6}$} & \cellcolor{gray!15}34.4\textcolor{myGreen}{$_{+7.6}$} & \cellcolor{gray!15}11.6\textcolor{myGreen}{$_{+3.2}$} \\
NSR & \cellcolor{gray!15}7.9\textcolor{myGreen}{$_{+4.6}$} & \cellcolor{gray!15}36.9\textcolor{myGreen}{$_{+4.4}$} & \cellcolor{gray!15}24.7\textcolor{myGreen}{$_{+4.5}$} & \cellcolor{gray!15}0.0\textcolor{myRed}{$_{-0.8}$} & \cellcolor{gray!15}34.2\textcolor{myRed}{$_{-4.4}$} & \cellcolor{gray!15}4.3\textcolor{myGreen}{$_{+0.2}$} & \cellcolor{green!15}67.2\textcolor{myGreen}{$_{+6.8}$} & \cellcolor{green!15}86.4\textcolor{myRed}{$_{0.0}$} & \cellcolor{gray!15}2.0\textcolor{myRed}{$_{0.0}$} & \cellcolor{gray!15}28.0\textcolor{myRed}{$_{-0.8}$} & \cellcolor{gray!15}5.2\textcolor{myRed}{$_{-3.2}$} \\
PSR & \cellcolor{gray!15}7.9\textcolor{myGreen}{$_{+4.6}$} & \cellcolor{gray!15}35.7\textcolor{myGreen}{$_{+4.2}$} & \cellcolor{gray!15}23.6\textcolor{myGreen}{$_{+3.4}$} & \cellcolor{gray!15}13.0\textcolor{myGreen}{$_{+11.5}$} & \cellcolor{gray!15}43.3\textcolor{myGreen}{$_{+4.7}$} & \cellcolor{gray!15}4.1\textcolor{myRed}{$_{0.0}$} & \cellcolor{green!15}69.2\textcolor{myGreen}{$_{+8.8}$} & \cellcolor{green!15}89.6\textcolor{myGreen}{$_{+3.2}$} & \cellcolor{gray!15}12.0\textcolor{myGreen}{$_{+11.2}$} & \cellcolor{gray!15}34.4\textcolor{myGreen}{$_{+7.6}$} & \cellcolor{gray!15}10.8\textcolor{myGreen}{$_{+2.4}$} \\
\bottomrule
\end{tabular}}
\caption{Results of NSR and PSR under different settings. When Model-Task alignment is strong, both NSR and PSR yield pronounced performance gains for all models (\textcolor{red}{Red} and \textcolor{green}{Green}). Conversely, under weak alignment, NSR-trained models exhibit no noticeable improvement (\textcolor{gray}{Gray}).
}
\label{tab:nsr}
\end{table}

%% file: 4-conclusion.tex
\section{Conclusion}
\label{sec:conc}
This work reveals that \emph{Model-Task Alignment} strength, measured by pass@k accuracy, serves as the fundamental determinant of when counterintuitive RL phenomena emerge in language model reasoning. We demonstrate that remarkable behaviors—including robustness to spurious rewards, one-shot training effectiveness, and negative-only signal sufficiency—manifest primarily when models already possess strong foundational capabilities in the target domain, functioning more as capability elicitation mechanisms rather than genuine learning drivers for unfamiliar tasks.

%% file: tables/more_contamination.tex
\begin{table}[!ht] 
\centering 
\footnotesize 
\renewcommand{\arraystretch}{1.3} 
\setlength{\tabcolsep}{4pt} 
\begin{tabular}{l l c cc cc cc cc} 
\toprule 
\multirow{2}{*}{\textbf{Task Type}} & \multirow{2}{*}{\textbf{Benchmark}} & \multirow{2}{*}{\textbf{Model}} & \multicolumn{2}{c}{\textbf{Portion=0.4}} & \multicolumn{2}{c}{\textbf{Portion=0.6}} & \multicolumn{2}{c}{\textbf{Portion=0.8}} \\ 
\cmidrule(lr){4-5} \cmidrule(lr){6-7} \cmidrule(lr){8-9} 
& & & \textbf{ROUGE} & \textbf{EM} & \textbf{ROUGE} & \textbf{EM} & \textbf{ROUGE} & \textbf{EM} \\ 
\midrule 
\multirow{6}{*}{\rotatebox{90}{\textbf{Math Tasks}}} 
& \multirow{2}{*}{\textbf{AMC 23}} & \textbf{Qwen2.5-7B} & 63.78\cellcolor{red!15} & 23.91\cellcolor{red!15} & 64.42\cellcolor{red!15} & 33.73\cellcolor{red!15} & 73.23\cellcolor{red!15} & 49.39\cellcolor{red!15} \\ 
& & \textbf{Llama-3.1-8B} & 27.18\cellcolor{gray!15} & 0.00\cellcolor{gray!15} & 30.64\cellcolor{gray!15} & 0.00\cellcolor{gray!15} & 44.54\cellcolor{gray!15} & 4.81\cellcolor{gray!15} \\ 
\cmidrule(lr){2-9} 
& \multirow{2}{*}{\textbf{MATH500}} & \textbf{Qwen2.5-7B} & 50.36\cellcolor{red!15} & 8.20\cellcolor{red!15} & 60.98\cellcolor{red!15} & 21.20\cellcolor{red!15} & 66.42\cellcolor{red!15} & 40.20\cellcolor{red!15} \\ 
& & \textbf{Llama-3.1-8B} & 23.09\cellcolor{gray!15} & 0.60\cellcolor{gray!15} & 40.56\cellcolor{gray!15} & 3.80\cellcolor{gray!15} & 48.33\cellcolor{gray!15} & 17.8\cellcolor{gray!15} \\ 
\cmidrule(lr){2-9} 
& \multirow{2}{*}{\textbf{AIME24}} & \textbf{Qwen2.5-7B} & 44.64\cellcolor{red!15} & 10.00\cellcolor{red!15} & 48.69\cellcolor{red!15} & 13.33\cellcolor{red!15} & 60.08\cellcolor{red!15} & 30.00\cellcolor{red!15} \\ 
& & \textbf{Llama-3.1-8B} & 26.08\cellcolor{gray!15} & 0.00\cellcolor{gray!15} & 30.80\cellcolor{gray!15} & 0.00\cellcolor{gray!15} & 50.50\cellcolor{gray!15} & 13.33\cellcolor{gray!15} \\ 
\midrule 
\multirow{12}{*}{\rotatebox{90}{\textbf{Logic Tasks}}} 
& \multirow{2}{*}{\textbf{Puzzle}} & \textbf{Qwen2.5-7B} & 19.56\cellcolor{gray!15} & 0.00\cellcolor{gray!15} & 19.62\cellcolor{gray!15} & 0.00\cellcolor{gray!15} & 19.24\cellcolor{gray!15} & 0.00\cellcolor{gray!15} \\ 
& & \textbf{Llama-3.1-8B} & 18.27\cellcolor{gray!15} & 0.00\cellcolor{gray!15} & 17.31\cellcolor{gray!15} & 0.00\cellcolor{gray!15} & 15.85\cellcolor{gray!15} & 0.00\cellcolor{gray!15} \\ 
\cmidrule(lr){2-9} 
& \multirow{2}{*}{\textbf{Operation}} & \textbf{Qwen2.5-7B} & 21.37\cellcolor{green!15} & 0.00\cellcolor{green!15} & 24.25\cellcolor{green!15} & 0.00\cellcolor{green!15} & 20.18\cellcolor{green!15} & 0.00\cellcolor{green!15} \\ 
& & \textbf{Llama-3.1-8B} & 21.83\cellcolor{green!15} & 0.00\cellcolor{green!15} & 18.34\cellcolor{green!15} & 0.00\cellcolor{green!15} & 16.75\cellcolor{green!15} & 0.00\cellcolor{green!15} \\ 
\cmidrule(lr){2-9} 
& \multirow{2}{*}{\textbf{Counterfactual}} & \textbf{Qwen2.5-7B} & 18.88\cellcolor{green!15} & 0.00\cellcolor{green!15} & 19.96\cellcolor{green!15} & 0.00\cellcolor{green!15} & 18.66\cellcolor{green!15} & 0.00\cellcolor{green!15} \\ 
& & \textbf{Llama-3.1-8B} & 19.02\cellcolor{green!15} & 0.00\cellcolor{green!15} & 19.39\cellcolor{green!15} & 0.00\cellcolor{green!15} & 18.94\cellcolor{green!15} & 0.00\cellcolor{green!15} \\ 
\cmidrule(lr){2-9} 
& \multirow{2}{*}{\textbf{Logic}} & \textbf{Qwen2.5-7B} & 22.08\cellcolor{gray!15} & 0.00\cellcolor{gray!15} & 27.28\cellcolor{gray!15} & 0.00\cellcolor{gray!15} & 28.23\cellcolor{gray!15} & 0.00\cellcolor{gray!15} \\ 
& & \textbf{Llama-3.1-8B} & 21.38\cellcolor{gray!15} & 0.00\cellcolor{gray!15} & 28.37\cellcolor{gray!15} & 0.00\cellcolor{gray!15} & 28.42\cellcolor{gray!15} & 0.00\cellcolor{gray!15} \\ 
\cmidrule(lr){2-9} 
& \multirow{2}{*}{\textbf{Cipher}} & \textbf{Qwen2.5-7B} & 34.61\cellcolor{gray!15} & 0.00\cellcolor{gray!15} & 41.03\cellcolor{gray!15} & 0.00\cellcolor{gray!15} & 44.77\cellcolor{gray!15} & 0.00\cellcolor{gray!15} \\ 
& & \textbf{Llama-3.1-8B} & 29.59\cellcolor{gray!15} & 0.00\cellcolor{gray!15} & 36.95\cellcolor{gray!15} & 0.00\cellcolor{gray!15} & 42.93\cellcolor{gray!15} & 0.00\cellcolor{gray!15} \\ 
\bottomrule 
\end{tabular} 
\caption{Extended Contamination Analysis across model-task combinations. \textcolor{red}{\textbf{Red}} \textcolor{red}{indicates potential contamination with strong baseline performance}; \textcolor{gray}{\textbf{Gray}} \textcolor{gray}{indicates no contamination with weak baseline performance}; \textcolor{green!70!black}{\textbf{Green}} \textcolor{green!70!black}{indicates no contamination with strong baseline performance.}} 
\label{tab:extended_contamination} 
\end{table}

%% file: tables/code.tex
\begin{table}[!ht]
    \begin{tabular}{@{}l cc cc@{}}
        \toprule
        \multirow{2.5}{*}{\textbf{Reward Type}} & \multicolumn{2}{c}{\textbf{MATH500}} & \multicolumn{2}{c}{\textbf{SynLogic}} \\
        \cmidrule(lr){2-3} \cmidrule(lr){4-5}
        & {\textbf{Before RL}} & {\textbf{After RL}} & {\textbf{Before RL}} & {\textbf{After RL}} \\
        \midrule
        Correct & \multirow{4}{*}{89.1} & \textbf{12.4} & \multirow{4}{*}{57.3} & \textbf{21.7} \\
        Random & & 94.2 & & 48.2 \\
        Format & & 96.7 & & 50.7 \\
        Incorrect & & 28.1 & & 28.3 \\
        \bottomrule
    \end{tabular}
    \caption{Code Usage Count of Qwen2.5-7B before and after RL training with different rewards.}
    \label{tab:code}
\end{table}

%% file: tables/code-llama.tex
\begin{table}[!ht]
    \begin{tabular}{@{}l cc cc@{}}
        \toprule
        \multirow{2.5}{*}{\textbf{Reward Type}} & \multicolumn{2}{c}{\textbf{Operation}} & \multicolumn{2}{c}{\textbf{Counterfactual}} \\
        \cmidrule(lr){2-3} \cmidrule(lr){4-5}
        & {\textbf{Before RL}} & {\textbf{After RL}} & {\textbf{Before RL}} & {\textbf{After RL}} \\
        \midrule
        Correct & \multirow{4}{*}{0.0} & 0.8 & \multirow{4}{*}{0.0} & 0.0 \\
        Random & & 0.0 & & 0.0 \\
        Format & & 0.0 & & 0.0 \\
        Incorrect & & 0.0 & & 0.0 \\
        \bottomrule
    \end{tabular}
    \caption{Code Usage Count of Llama-3.1-8B-Instruct before and after RL training on two tasks.}
    \label{tab:code-llama}
\end{table}

%% file: 0-main.bbl
\begin{thebibliography}{34}
\providecommand{\natexlab}[1]{#1}
\providecommand{\url}[1]{\texttt{#1}}
\expandafter\ifx\csname urlstyle\endcsname\relax
  \providecommand{\doi}[1]{doi: #1}\else
  \providecommand{\doi}{doi: \begingroup \urlstyle{rm}\Url}\fi

\bibitem[Agarwal et~al.(2025)Agarwal, Zhang, Yuan, Han, and Peng]{agarwal2025unreasonableeffectivenessentropyminimization}
Shivam Agarwal, Zimin Zhang, Lifan Yuan, Jiawei Han, and Hao Peng.
\newblock The unreasonable effectiveness of entropy minimization in llm reasoning, 2025.
\newblock URL \url{https://arxiv.org/abs/2505.15134}.

\bibitem[AIME(2024)]{aime24}
AIME.
\newblock {A}rt of {P}roblem {S}olving --- artofproblemsolving.com.
\newblock \url{https://artofproblemsolving.com/wiki/index.php/AIME_Problems_and_Solutions}, 2024.
\newblock [Accessed 26-08-2025].

\bibitem[AMC(2023)]{amc23}
AMC.
\newblock {A}rt of {P}roblem {S}olving --- artofproblemsolving.com.
\newblock \url{https://artofproblemsolving.com/wiki/index.php/AMC_12_Problems_and_Solutions}, 2023.
\newblock [Accessed 26-08-2025].

\bibitem[Chen et~al.(2025)Chen, He, Yuan, Chen, Cai, Dai, Yu, Yu, Li, Chen, Zhou, and Wang]{chen2025enigmatascalinglogicalreasoning}
Jiangjie Chen, Qianyu He, Siyu Yuan, Aili Chen, Zhicheng Cai, Weinan Dai, Hongli Yu, Qiying Yu, Xuefeng Li, Jiaze Chen, Hao Zhou, and Mingxuan Wang.
\newblock Enigmata: Scaling logical reasoning in large language models with synthetic verifiable puzzles, 2025.
\newblock URL \url{https://arxiv.org/abs/2505.19914}.

\bibitem[Chen et~al.(2024)Chen, Zhu, Sun, Chen, Zhang, and Shen]{chen2024accuracy}
Yanjun Chen, Dawei Zhu, Yirong Sun, Xinghao Chen, Wei Zhang, and Xiaoyu Shen.
\newblock The accuracy paradox in rlhf: When better reward models don't yield better language models.
\newblock \emph{arXiv preprint arXiv:2410.06554}, 2024.

\bibitem[Guo et~al.(2025)Guo, Yang, Zhang, Song, Zhang, Xu, Zhu, Ma, Wang, Bi, et~al.]{deepseekr1}
Daya Guo, Dejian Yang, Haowei Zhang, Junxiao Song, Ruoyu Zhang, Runxin Xu, Qihao Zhu, Shirong Ma, Peiyi Wang, Xiao Bi, et~al.
\newblock Deepseek-r1: Incentivizing reasoning capability in llms via reinforcement learning.
\newblock \emph{arXiv preprint arXiv:2501.12948}, 2025.

\bibitem[He et~al.(2024)He, Luo, Bai, Hu, Thai, Shen, Hu, Han, Huang, Zhang, Liu, Qi, Liu, and Sun]{he2024olympiadbench}
Chaoqun He, Renjie Luo, Yuzhuo Bai, Shengding Hu, Zhen~Leng Thai, Junhao Shen, Jinyi Hu, Xu~Han, Yujie Huang, Yuxiang Zhang, Jie Liu, Lei Qi, Zhiyuan Liu, and Maosong Sun.
\newblock Olympiadbench: A challenging benchmark for promoting agi with olympiad-level bilingual multimodal scientific problems, 2024.

\bibitem[Hendrycks et~al.(2021)Hendrycks, Burns, Kadavath, Arora, Basart, Tang, Song, and Steinhardt]{hendrycks2021measuringmathematicalproblemsolving}
Dan Hendrycks, Collin Burns, Saurav Kadavath, Akul Arora, Steven Basart, Eric Tang, Dawn Song, and Jacob Steinhardt.
\newblock Measuring mathematical problem solving with the math dataset, 2021.
\newblock URL \url{https://arxiv.org/abs/2103.03874}.

\bibitem[Jaech et~al.(2024)Jaech, Kalai, Lerer, Richardson, El-Kishky, Low, Helyar, Madry, Beutel, Carney, et~al.]{openaio1}
Aaron Jaech, Adam Kalai, Adam Lerer, Adam Richardson, Ahmed El-Kishky, Aiden Low, Alec Helyar, Aleksander Madry, Alex Beutel, Alex Carney, et~al.
\newblock Openai o1 system card.
\newblock \emph{arXiv preprint arXiv:2412.16720}, 2024.

\bibitem[Jain et~al.(2024)Jain, Han, Gu, Li, Yan, Zhang, Wang, Solar-Lezama, Sen, and Stoica]{jain2024livecodebenchholisticcontaminationfree}
Naman Jain, King Han, Alex Gu, Wen-Ding Li, Fanjia Yan, Tianjun Zhang, Sida Wang, Armando Solar-Lezama, Koushik Sen, and Ion Stoica.
\newblock Livecodebench: Holistic and contamination free evaluation of large language models for code, 2024.
\newblock URL \url{https://arxiv.org/abs/2403.07974}.

\bibitem[Kazemi et~al.(2025)Kazemi, Fatemi, Bansal, Palowitch, Anastasiou, Mehta, Jain, Aglietti, Jindal, Chen, Dikkala, Tyen, Liu, Shalit, Chiappa, Olszewska, Tay, Tran, Le, and Firat]{bbeh}
Mehran Kazemi, Bahare Fatemi, Hritik Bansal, John Palowitch, Chrysovalantis Anastasiou, Sanket~Vaibhav Mehta, Lalit~K. Jain, Virginia Aglietti, Disha Jindal, Peter Chen, Nishanth Dikkala, Gladys Tyen, Xin Liu, Uri Shalit, Silvia Chiappa, Kate Olszewska, Yi~Tay, Vinh~Q. Tran, Quoc~V. Le, and Orhan Firat.
\newblock Big-bench extra hard, 2025.
\newblock URL \url{https://arxiv.org/abs/2502.19187}.

\bibitem[Lewkowycz et~al.(2022)Lewkowycz, Andreassen, Dohan, Dyer, Michalewski, Ramasesh, Slone, Anil, Schlag, Gutman-Solo, et~al.]{minervamath}
Aitor Lewkowycz, Anders Andreassen, David Dohan, Ethan Dyer, Henryk Michalewski, Vinay Ramasesh, Ambrose Slone, Cem Anil, Imanol Schlag, Theo Gutman-Solo, et~al.
\newblock Solving quantitative reasoning problems with language models.
\newblock \emph{Advances in Neural Information Processing Systems}, 35:\penalty0 3843--3857, 2022.

\bibitem[Liu et~al.(2025)Liu, Fan, Jiang, Ding, Hu, Zhang, Shi, Weng, Chen, Chen, Huang, Zhang, Zhao, Yan, and He]{liu2025synlogic}
Junteng Liu, Yuanxiang Fan, Zhuo Jiang, Han Ding, Yongyi Hu, Chi Zhang, Yiqi Shi, Shitong Weng, Aili Chen, Shiqi Chen, Yunan Huang, Mozhi Zhang, Pengyu Zhao, Junjie Yan, and Junxian He.
\newblock Synlogic: Synthesizing verifiable reasoning data at scale for learning logical reasoning and beyond, 2025.
\newblock URL \url{https://arxiv.org/abs/2505.19641}.

\bibitem[Luo et~al.(2025)Luo, Tan, Wong, Shi, Tang, Roongta, Cai, Luo, Zhang, Li, Popa, and Stoica]{deepscaler2025}
Michael Luo, Sijun Tan, Justin Wong, Xiaoxiang Shi, William Tang, Manan Roongta, Colin Cai, Jeffrey Luo, Tianjun Zhang, Erran Li, Raluca~Ada Popa, and Ion Stoica.
\newblock Deepscaler: Surpassing o1-preview with a 1.5b model by scaling rl.
\newblock \href{https://pretty-radio-b75.notion.site/DeepScaleR-Surpassing-O1-Preview-with-a-1-5B-Model-by-Scaling-RL-19681902c1468005bed8ca303013a4e2}{Link}, 2025.
\newblock Notion Blog.

\bibitem[Lv et~al.(2025)Lv, Xie, Sun, Kang, and Yan]{lv2025climb}
Ang Lv, Ruobing Xie, Xingwu Sun, Zhanhui Kang, and Rui Yan.
\newblock The climb carves wisdom deeper than the summit: On the noisy rewards in learning to reason.
\newblock \emph{arXiv preprint arXiv:2505.22653}, 2025.

\bibitem[Ma et~al.(2024)Ma, Du, Wang, Zhang, Wen, Qu, Yang, Liu, Liu, Yue, Huang, and Zhang]{ma2024korbenchbenchmarkinglanguagemodels}
Kaijing Ma, Xinrun Du, Yunran Wang, Haoran Zhang, Zhoufutu Wen, Xingwei Qu, Jian Yang, Jiaheng Liu, Minghao Liu, Xiang Yue, Wenhao Huang, and Ge~Zhang.
\newblock Kor-bench: Benchmarking language models on knowledge-orthogonal reasoning tasks, 2024.
\newblock URL \url{https://arxiv.org/abs/2410.06526}.

\bibitem[Meta(2024)]{llama3}
Meta.
\newblock The llama 3 herd of models, 2024.
\newblock URL \url{https://arxiv.org/abs/2407.21783}.

\bibitem[Ouyang et~al.(2022)Ouyang, Wu, Jiang, Almeida, Wainwright, Mishkin, Zhang, Agarwal, Slama, Ray, et~al.]{ouyang2022training}
Long Ouyang, Jeffrey Wu, Xu~Jiang, Diogo Almeida, Carroll Wainwright, Pamela Mishkin, Chong Zhang, Sandhini Agarwal, Katarina Slama, Alex Ray, et~al.
\newblock Training language models to follow instructions with human feedback.
\newblock \emph{Advances in neural information processing systems}, 35:\penalty0 27730--27744, 2022.

\bibitem[Qwen et~al.(2025)Qwen, :, Yang, Yang, Zhang, Hui, Zheng, Yu, Li, Liu, Huang, Wei, Lin, Yang, Tu, Zhang, Yang, Yang, Zhou, Lin, Dang, Lu, Bao, Yang, Yu, Li, Xue, Zhang, Zhu, Men, Lin, Li, Tang, Xia, Ren, Ren, Fan, Su, Zhang, Wan, Liu, Cui, Zhang, and Qiu]{qwen2.5}
Qwen, :, An~Yang, Baosong Yang, Beichen Zhang, Binyuan Hui, Bo~Zheng, Bowen Yu, Chengyuan Li, Dayiheng Liu, Fei Huang, Haoran Wei, Huan Lin, Jian Yang, Jianhong Tu, Jianwei Zhang, Jianxin Yang, Jiaxi Yang, Jingren Zhou, Junyang Lin, Kai Dang, Keming Lu, Keqin Bao, Kexin Yang, Le~Yu, Mei Li, Mingfeng Xue, Pei Zhang, Qin Zhu, Rui Men, Runji Lin, Tianhao Li, Tianyi Tang, Tingyu Xia, Xingzhang Ren, Xuancheng Ren, Yang Fan, Yang Su, Yichang Zhang, Yu~Wan, Yuqiong Liu, Zeyu Cui, Zhenru Zhang, and Zihan Qiu.
\newblock Qwen2.5 technical report, 2025.
\newblock URL \url{https://arxiv.org/abs/2412.15115}.

\bibitem[Rafailov et~al.(2024)Rafailov, Sharma, Mitchell, Ermon, Manning, and Finn]{rafailov2024directpreferenceoptimizationlanguage}
Rafael Rafailov, Archit Sharma, Eric Mitchell, Stefano Ermon, Christopher~D. Manning, and Chelsea Finn.
\newblock Direct preference optimization: Your language model is secretly a reward model, 2024.
\newblock URL \url{https://arxiv.org/abs/2305.18290}.

\bibitem[Shao et~al.(2025)Shao, Li, Xin, Geng, Wang, Oh, Du, Lambert, Min, Krishna, et~al.]{shao2025spurious}
Rulin Shao, Shuyue~Stella Li, Rui Xin, Scott Geng, Yiping Wang, Sewoong Oh, Simon~Shaolei Du, Nathan Lambert, Sewon Min, Ranjay Krishna, et~al.
\newblock Spurious rewards: Rethinking training signals in rlvr.
\newblock \emph{arXiv preprint arXiv:2506.10947}, 2025.

\bibitem[Silver et~al.(2017{\natexlab{a}})Silver, Hubert, Schrittwieser, Antonoglou, Lai, Guez, Lanctot, Sifre, Kumaran, Graepel, Lillicrap, Simonyan, and Hassabis]{silver2017masteringchessshogiselfplay}
David Silver, Thomas Hubert, Julian Schrittwieser, Ioannis Antonoglou, Matthew Lai, Arthur Guez, Marc Lanctot, Laurent Sifre, Dharshan Kumaran, Thore Graepel, Timothy Lillicrap, Karen Simonyan, and Demis Hassabis.
\newblock Mastering chess and shogi by self-play with a general reinforcement learning algorithm, 2017{\natexlab{a}}.
\newblock URL \url{https://arxiv.org/abs/1712.01815}.

\bibitem[Silver et~al.(2017{\natexlab{b}})Silver, Schrittwieser, Simonyan, Antonoglou, Huang, Guez, Hubert, baker, Lai, Bolton, Chen, Lillicrap, Hui, Sifre, van~den Driessche, Graepel, and Hassabis]{Silver2017MasteringT}
David Silver, Julian Schrittwieser, Karen Simonyan, Ioannis Antonoglou, Aja Huang, Arthur Guez, Thomas Hubert, Lucas baker, Matthew Lai, Adrian Bolton, Yutian Chen, Timothy~P. Lillicrap, Fan Hui, L.~Sifre, George van~den Driessche, Thore Graepel, and Demis Hassabis.
\newblock Mastering the game of go without human knowledge.
\newblock \emph{Nature}, 550:\penalty0 354--359, 2017{\natexlab{b}}.
\newblock URL \url{https://api.semanticscholar.org/CorpusID:205261034}.

\bibitem[Sutton et~al.(1998)Sutton, Barto, et~al.]{sutton1998reinforcement}
Richard~S Sutton, Andrew~G Barto, et~al.
\newblock \emph{Reinforcement learning: An introduction}, volume~1.
\newblock MIT press Cambridge, 1998.

\bibitem[Suzgun et~al.(2022)Suzgun, Scales, Sch{\"a}rli, Gehrmann, Tay, Chung, Chowdhery, Le, Chi, Zhou, , and Wei]{suzgun2022challenging}
Mirac Suzgun, Nathan Scales, Nathanael Sch{\"a}rli, Sebastian Gehrmann, Yi~Tay, Hyung~Won Chung, Aakanksha Chowdhery, Quoc~V Le, Ed~H Chi, Denny Zhou, , and Jason Wei.
\newblock Challenging big-bench tasks and whether chain-of-thought can solve them.
\newblock \emph{arXiv preprint arXiv:2210.09261}, 2022.

\bibitem[Team et~al.(2025)Team, Du, Gao, Xing, Jiang, Chen, Li, Xiao, Du, Liao, et~al.]{team2025kimi}
Kimi Team, Angang Du, Bofei Gao, Bowei Xing, Changjiu Jiang, Cheng Chen, Cheng Li, Chenjun Xiao, Chenzhuang Du, Chonghua Liao, et~al.
\newblock Kimi k1. 5: Scaling reinforcement learning with llms.
\newblock \emph{arXiv preprint arXiv:2501.12599}, 2025.

\bibitem[Team(2025)]{qwq32b}
Qwen Team.
\newblock Qwq-32b: Embracing the power of reinforcement learning, March 2025.
\newblock URL \url{https://qwenlm.github.io/blog/qwq-32b/}.

\bibitem[Wang et~al.(2025)Wang, Yang, Zeng, Ren, Liu, Peng, Cheng, He, Wang, Gao, et~al.]{wang2025reinforcement}
Yiping Wang, Qing Yang, Zhiyuan Zeng, Liliang Ren, Liyuan Liu, Baolin Peng, Hao Cheng, Xuehai He, Kuan Wang, Jianfeng Gao, et~al.
\newblock Reinforcement learning for reasoning in large language models with one training example.
\newblock \emph{arXiv preprint arXiv:2504.20571}, 2025.

\bibitem[Wu et~al.(2025)Wu, Zhang, Dong, Xi, Zhao, Jin, Fan, Zhou, Fu, Liu, et~al.]{wu2025reasoning}
Mingqi Wu, Zhihao Zhang, Qiaole Dong, Zhiheng Xi, Jun Zhao, Senjie Jin, Xiaoran Fan, Yuhao Zhou, Yanwei Fu, Qin Liu, et~al.
\newblock Reasoning or memorization? unreliable results of reinforcement learning due to data contamination.
\newblock \emph{arXiv preprint arXiv:2507.10532}, 2025.

\bibitem[Yang et~al.(2024)Yang, Zhang, Hui, Gao, Yu, Li, Liu, Tu, Zhou, Lin, Lu, Xue, Lin, Liu, Ren, and Zhang]{yang2024qwen25math}
An~Yang, Beichen Zhang, Binyuan Hui, Bofei Gao, Bowen Yu, Chengpeng Li, Dayiheng Liu, Jianhong Tu, Jingren Zhou, Junyang Lin, Keming Lu, Mingfeng Xue, Runji Lin, Tianyu Liu, Xingzhang Ren, and Zhenru Zhang.
\newblock Qwen2.5-math technical report: Toward mathematical expert model via self-improvement, 2024.
\newblock URL \url{https://arxiv.org/abs/2409.12122}.

\bibitem[Yang et~al.(2025)Yang, Li, Yang, Zhang, Hui, Zheng, Yu, Gao, Huang, Lv, Zheng, Liu, Zhou, Huang, Hu, Ge, Wei, Lin, Tang, Yang, Tu, Zhang, Yang, Yang, Zhou, Zhou, Lin, Dang, Bao, Yang, Yu, Deng, Li, Xue, Li, Zhang, Wang, Zhu, Men, Gao, Liu, Luo, Li, Tang, Yin, Ren, Wang, Zhang, Ren, Fan, Su, Zhang, Zhang, Wan, Liu, Wang, Cui, Zhang, Zhou, and Qiu]{yang2025qwen3technicalreport}
An~Yang, Anfeng Li, Baosong Yang, Beichen Zhang, Binyuan Hui, Bo~Zheng, Bowen Yu, Chang Gao, Chengen Huang, Chenxu Lv, Chujie Zheng, Dayiheng Liu, Fan Zhou, Fei Huang, Feng Hu, Hao Ge, Haoran Wei, Huan Lin, Jialong Tang, Jian Yang, Jianhong Tu, Jianwei Zhang, Jianxin Yang, Jiaxi Yang, Jing Zhou, Jingren Zhou, Junyang Lin, Kai Dang, Keqin Bao, Kexin Yang, Le~Yu, Lianghao Deng, Mei Li, Mingfeng Xue, Mingze Li, Pei Zhang, Peng Wang, Qin Zhu, Rui Men, Ruize Gao, Shixuan Liu, Shuang Luo, Tianhao Li, Tianyi Tang, Wenbiao Yin, Xingzhang Ren, Xinyu Wang, Xinyu Zhang, Xuancheng Ren, Yang Fan, Yang Su, Yichang Zhang, Yinger Zhang, Yu~Wan, Yuqiong Liu, Zekun Wang, Zeyu Cui, Zhenru Zhang, Zhipeng Zhou, and Zihan Qiu.
\newblock Qwen3 technical report, 2025.
\newblock URL \url{https://arxiv.org/abs/2505.09388}.

\bibitem[Zhu et~al.(2025)Zhu, Xia, Wei, Chen, Chen, and Meng]{zhu2025surprisingeffectivenessnegativereinforcement}
Xinyu Zhu, Mengzhou Xia, Zhepei Wei, Wei-Lin Chen, Danqi Chen, and Yu~Meng.
\newblock The surprising effectiveness of negative reinforcement in llm reasoning, 2025.
\newblock URL \url{https://arxiv.org/abs/2506.01347}.

\bibitem[Ziegler et~al.(2019)Ziegler, Stiennon, Wu, Brown, Radford, Amodei, Christiano, and Irving]{ziegler2019fine}
Daniel~M Ziegler, Nisan Stiennon, Jeffrey Wu, Tom~B Brown, Alec Radford, Dario Amodei, Paul Christiano, and Geoffrey Irving.
\newblock Fine-tuning language models from human preferences.
\newblock \emph{arXiv preprint arXiv:1909.08593}, 2019.

\bibitem[Zuo et~al.(2025)Zuo, Zhang, Sheng, Qu, Cui, Zhu, Li, Zhang, Long, Hua, et~al.]{zuo2025ttrl}
Yuxin Zuo, Kaiyan Zhang, Li~Sheng, Shang Qu, Ganqu Cui, Xuekai Zhu, Haozhan Li, Yuchen Zhang, Xinwei Long, Ermo Hua, et~al.
\newblock Ttrl: Test-time reinforcement learning.
\newblock \emph{arXiv preprint arXiv:2504.16084}, 2025.

\end{thebibliography}
